\documentclass[journal]{IEEEtran}
\IEEEoverridecommandlockouts
\pdfoutput=1
\usepackage{cite}
\usepackage{units}
\usepackage{lipsum}
\usepackage[caption=false]{subfig}
\usepackage{graphicx}	
\usepackage{pstricks}
\usepackage{float}
\usepackage{eurosym}
\usepackage{multicol}
\usepackage{booktabs}
\usepackage{units}
\usepackage{amsmath,amssymb}
\usepackage{pifont}

\graphicspath{{Figure/}}
\begin{document}

\title{MPC path-planner for autonomous driving solved by genetic algorithm technique}


\author{\IEEEauthorblockN{S. Arrigoni, F. Braghin, F. Cheli}
\IEEEauthorblockA{Politecnico di Milano, Mechanical Department\\
via G. La Masa, 1 - 20156 Milano Italy\\
Email: stefano.arrigoni@polimi.it\\
}

}


%


\maketitle

\begin{abstract}

Autonomous vehicle's technology is expected to be disruptive for automotive industry in next years. This paper proposes a novel real-time trajectory planner based on a Nonlinear Model Predictive Control (NMPC) algorithm.
A nonlinear single track vehicle model with Pacejka’s lateral tyre formulas has been implemented. 
The numerical solution of the NMPC problem is obtained by means of the implementation of a novel genetic algorithm strategy.
Numerical results are discussed through simulations that shown a reasonable behavior of the proposed strategy in presence of static or moving obstacles as well as in a wide rage of road friction conditions. Moreover a real-time implementation is made possible by the reported computational time analysis.
\end{abstract}


%
\IEEEpeerreviewmaketitle

\section{introduction}
Autonomous vehicle appeal grown exponentially in last years and it is expected to be disruptive for automotive industry involving not only traditional OEMs, but also suppliers, technology providers, academic institutions, municipal governments, and regulatory bodies. 
Considering that worldwide every year dies around 1.35 million people due to road accidents caused mainly by wrong behavior of the drivers as distracted driving, exceeding speed limits and driving under the influence of alcohol or drugs among the others\cite{WHO_18}, Autonomous driving that aims at reducing human control on driving tasks, could be considered a solution to increase road safety.
Apart from safety, Autonomous driving will lead to several possible advantages: pollution, traffic, space consumption and mobility cost reduction just to cite a few as reported by \cite{BSC_jul16}.

As suggested in \cite{BSC_apr15}, Autonomous Vehicle's development not only involves technological aspects, but it's intrinsically connected to societal, legal, and regulatory issues related to them and analysts (and most industry players and experts agrees on that) predicts that "The car of the future is electrified, autonomous, shared, connected and yearly updated" \cite{PWC_18}.

The most of industrial carmakers already revealed their pilot programs and prototypes of accurate localization and environment perception, highly automated or fully autonomous vehicles.

Significant examples are Daimler's Bertha prototype that autonomously revisited the "Bertha Benz Memorial Route" 125 years after Bertha Benz's first cross-country automobile journey in history \cite{bertha_1} as well as Tesla, leading company in term of media exposure in ADAS development, that already commercialized an highly automated semi-autonomous driving system called "autopilot" mounted on it's vehicles expected to receive incremental improvements in term of additional functionalities in next years.

Academia involvement in this field is well documented in literature. Just to cite a few: Stanford has several ongoing projects on autonomous vehicle prototypes. In \cite{stanf_1} the so-called VW Passat prototype "Stanley", which participated the DARPA Grand Challenge of 2005, is described and in \cite{audi1} the Audi research vehicle "Shelley" is presented. MIT worked on "Talos" \cite{MIT1} for DARPA Urban Challenge 2007. Vislab (Parma University) developed several prototypes over last decades obtaining impressive results during VisLab Intercontinental Autonomous Challenge (VIAC)\cite{Vislab1}. 

This paper, as most of the works in literature, is based on a hierarchical control structure composed by different layers to decouple the complete driving problem into a sequence of tasks \cite{survey_frazzoli,scheme_survey}.

In this paper, motion planning task is mainly investigated. In literature, this problem is tackled by different techniques as graph search methods which generate a path by means of an optimization of an objective function defined on nodes and edges in the discretized configuration space of the problem, A-star is an improved version of the famous “Dijkstra’s algorithm” obtained adding an heuristic term in objective function that estimates the cost of the cheapest path from a given node to the target. It is proven that A-star is usually faster than Dijkstra and in the worst case they have te same efficiency. In \cite{survey_mot2013,survey_frazzoli} some example are provided and $A^{\ast}$ extended versions are grouped into sub-families related to the innovative aspects improved.

Incremental search methods iteratively increment a graph of feasible configurations and transitions in the C-space until the target region is reached. These methods try to remove the dependence of the solution from the discretization of configuration space of the problem. Among several implementation proposed, like i.e. expansive spaces tree (EST) \cite{expansivespace}, Rapidly-exploring Random Tree (RRT) technique \cite{Lavalle98} rapidly spread and gained importance becoming one of the most used methods thanks to its high efficiency in handling high-dimensional non-holonomic problems. RRT$^{\ast}$, presented in \cite{Frazzoli2010}, is an "asymptotically optimal" improved version of the original RRT producing an increasingly finer discretization of the configuration space.

Potential field approach can be easily explained by means of an electric charge moving in a potential field. Firstly introduced by \cite{khatib}, it gained popularity for its simple implementation and extremely low computational demanding suitable for real-time application. Several improved implementation were studied during the years in order to solve local solution convergence problem as reported in the survey \cite{PF_survery} and it's nowadays an extremely interesting and wide spread on-line path planning technique.

Model Predictive Control (MPC) is an important tool for optimal control problem. Thanks to progress in term of mathematical formulation and a progressive increment of computational power available, it is possible to extend its application to real time motion planning and tracking tasks \cite{liniger2015optimization,frasch2013auto}.

MPC formulation has some interesting features that determined its increasing success among other techniques for complex control problems: it can handle a multivariable process subjected to inequality constraints on input and state variables, high delays and highly nonlinear systems; control strategy can vary dynamically in time in order to take into account changes in working condition (model, constraints or cost terms).

On the other hand, the main drawback of MPC is an high computational demand for its implementation (at every time step the entire optimization problem needs to be solved). 
Moreover the method relies on the open-loop integration of system dynamic over the horizon: this means that performances are highly dependent by model accuracy and reliability (i.e. vehicle model and obstacles' motion prediction on the road).

Examples available in literature mainly differ in term of vehicle model used, mathematical formulation of obstacles in the optimization problem and mathematical implementation of numerical solving method for the constrained optimization problem.

A common approach adopted in literature to reduce overall complexity, is to split trajectory generation and tracking tasks into a hierarchical structure composed by subsequent MPCs, where higher level aims to define optimal trajectory over a long time horizon based on a simplified model, while lower control level is in charge of tracking it by means of a more sophisticated vehicle model and a shorted time horizon.

As an example in \cite{gao2010predictive} high level controller is designed as a NMPC based on a point mass model capable of deal with static obstacle on a slippery road, and a lower level control based on a double track vehicle model coupled with Pacejka's tyre model. The overall resulting controller is real-time but the high discrepancy between the models used are leading to high tracking errors.
An improvement is presented in \cite{gray2012predictive}, where higher level control is modifed as a NMPC based on precomputed motion primitives based on a complex vehicle model. The new approach allows a reduction on tracking error but limit the solution to a limited set of suboptimal trajectories and requires an online implementation of the mixed-integer problem that can be difficult when a large number of precomputed maneuvers is considered.

As an improvement, in \cite{gao2012spatial} a two-level NMPC with a High level base on a nonlinear dynamic single-track vehicle model defined by mean of a spatial reformulation that considers a simplified Pacejka's tyre model is presented. this formulation allows to speed up calculation in presence of static obstacle and the more sophisticated tyre modeling improved reference trajectory quality reducing low level tracking error. The same formulation was used in \cite{frasch2013auto} where implementing a single level NMPC based on a complex model (four wheels model 6 dof with Pacejka's tyre model) merging trajectory generation and tracking tasks, tracking issue has been solved. The result was tested in simulation with static obstacles on low friction surfaces and impleented in real-time.
Unfortunately spatial reformulation presents important limitations as inability to manage velocity equal to zero and dramatically reduces its advantages in presence of moving obstacles.

Another dynamic single stage implementation is reported in \cite{liniger2015optimization} where single-track vehicle model with Pacejka's lateral tyre model is adopted. Calculation is speed up thanks to linearization around current configuration and considering moving obstacles as static within every MPC calculation.
An example of a kinematic single layer implementation is reported in \cite{gutjahr2017lateral} in the form of a Linear Time-Varying MPC where static and moving obstacles are considered, and the optimal control problem (OCP) is solved by using a quadratic programming (QP) routine.

Vehicle modeling problem for MPC applications is investigated in \cite{kong2015kinematic} where, after highlighting kinematic models advantages as lower computational effort and the possibility to handle stopped vehicles (singularity at zero velocity is related to slip calculation affecting dynamic tyre models), limitations as impossibility to consider friction limits at tyre-road interaction and modeling errors brings the authors to suggest using dynamic models in presence of moderate driving speeds.

The aim of this work is to propose a single layer NMPC formulation for trajectory generation and tracking problems with a special focus on real-time capability and robustness of the algorithm in a urban-like scenarios.
The vehicle model considered is a single-track vehicle model where lateral tyre dynamics are modelled through Pacejka's simplified Magic Formula. The algorithm is capable of handling multiple moving obstacles.
The numerical solution of the OCP is obtained by means of the implementation of a novel genetic algorithm strategy described in the paper.
For testing purposes, the resulting algorithm has been implemented in a Simulink simulation environment where robustness to model uncertainties and obstacle avoidance has been tested.

Section~\ref{sec:Hier} introduces assumptions and aim of the paper, in particular the hierarchical approach adopted to define the decisional algorithm involving trajectory planning and vehicle control.
In Section~\ref{sec:PP} the optimization problem is deeply described and the MPC formulation considered is presented.
Section~\ref{sec:GA} describes the genetic algorithm strategy adopted in order to solve the problem previously presented.
In section~\ref{sec:res} numerical simulations are shown in order to evaluate the behavior of the algorithm in term of performances, robustness and computational time required
Finally in section~\ref{sec:CONC} the conclusion are drawn.

\section{Assumptions and aim of the paper}\label{sec:Hier}

\begin{figure}[htp]\centering
	\includegraphics[width=0.95\columnwidth]{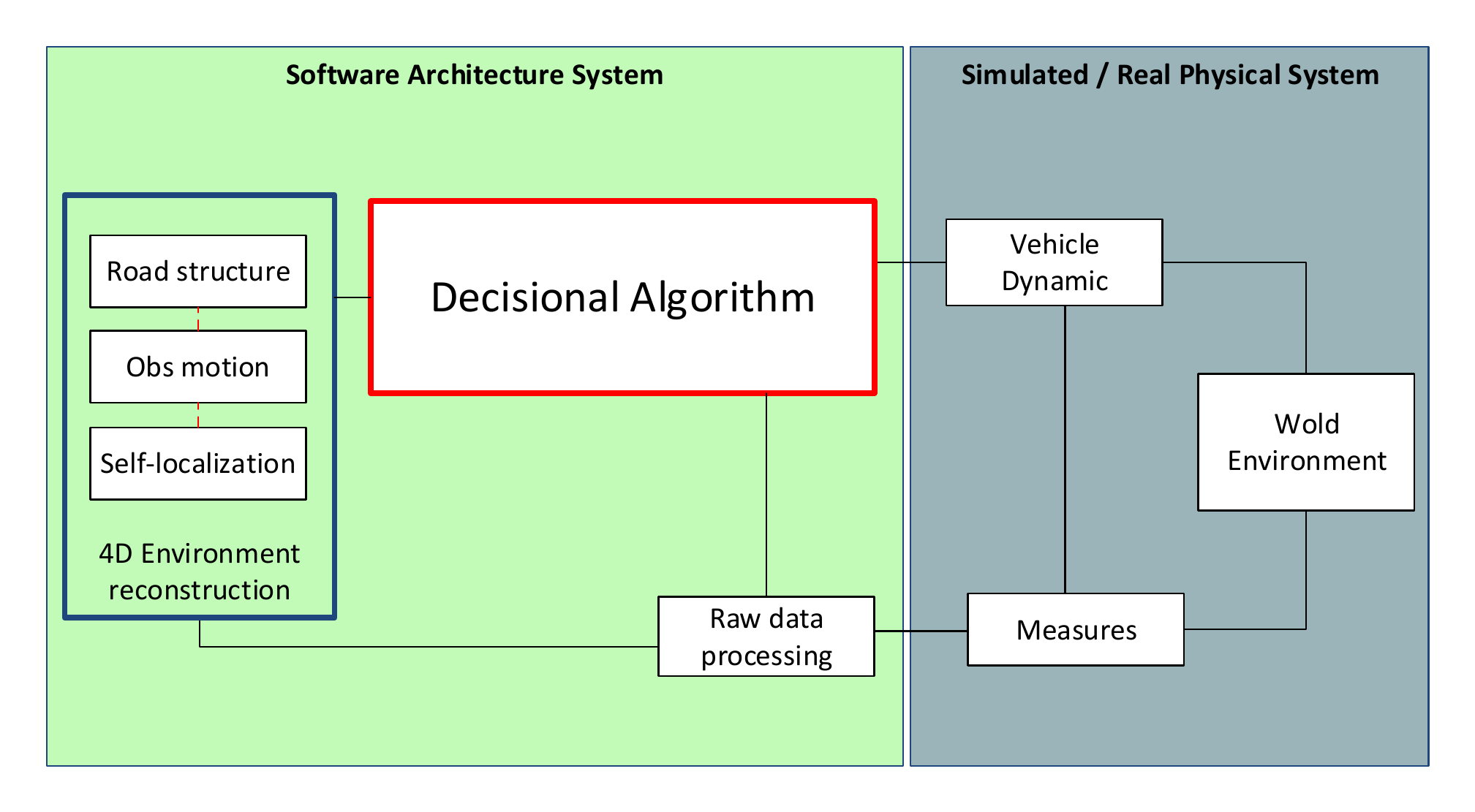}
	\caption{sub-tasks for Autonomous driving}
	\label{fig:general_scheme}
\end{figure}

According to prototype and survey papers \cite{junior,scheme_survey,annieway} available in literature, fig. \ref{fig:general_scheme} reports a general framework grouping all the main sub-tasks required for autonomous driving development.

In detail, on the right (blue background) physical (or simulated) system and environment are represented: it represents the "controlled" vehicle surrounded by the scenario (roads, other vehicles or pedestrians, \dots) and sensor setup.

On the left (green background) software algortihms are represented.
It can be mainly divided into three main sequential tasks:

\begin{itemize}
	\item "Raw data processing", where all data directly provided by sensors need to be processed in order to compute useful information relevant for decision making process or environment reconstruction;
	\item "4D environment reconstruction". Starting from processed data it's essential to reconstruct surrounding of the vehicle using redundant measures. In detail: accurate and reliable self-localization of the vehicle in a 3D digital map enriched by traffic and behavior laws, where obstacles are detected, tracked and their time evolution is predicted (time dimension); 
	\item "Decisional Algorithm"  groups all decisional driving tasks. starting from environment reconstruction and driver requests, it's aim is to define and actuate a decisional strategy.
\end{itemize}

\begin{figure}[htp]\centering
	\includegraphics[width=1\columnwidth]{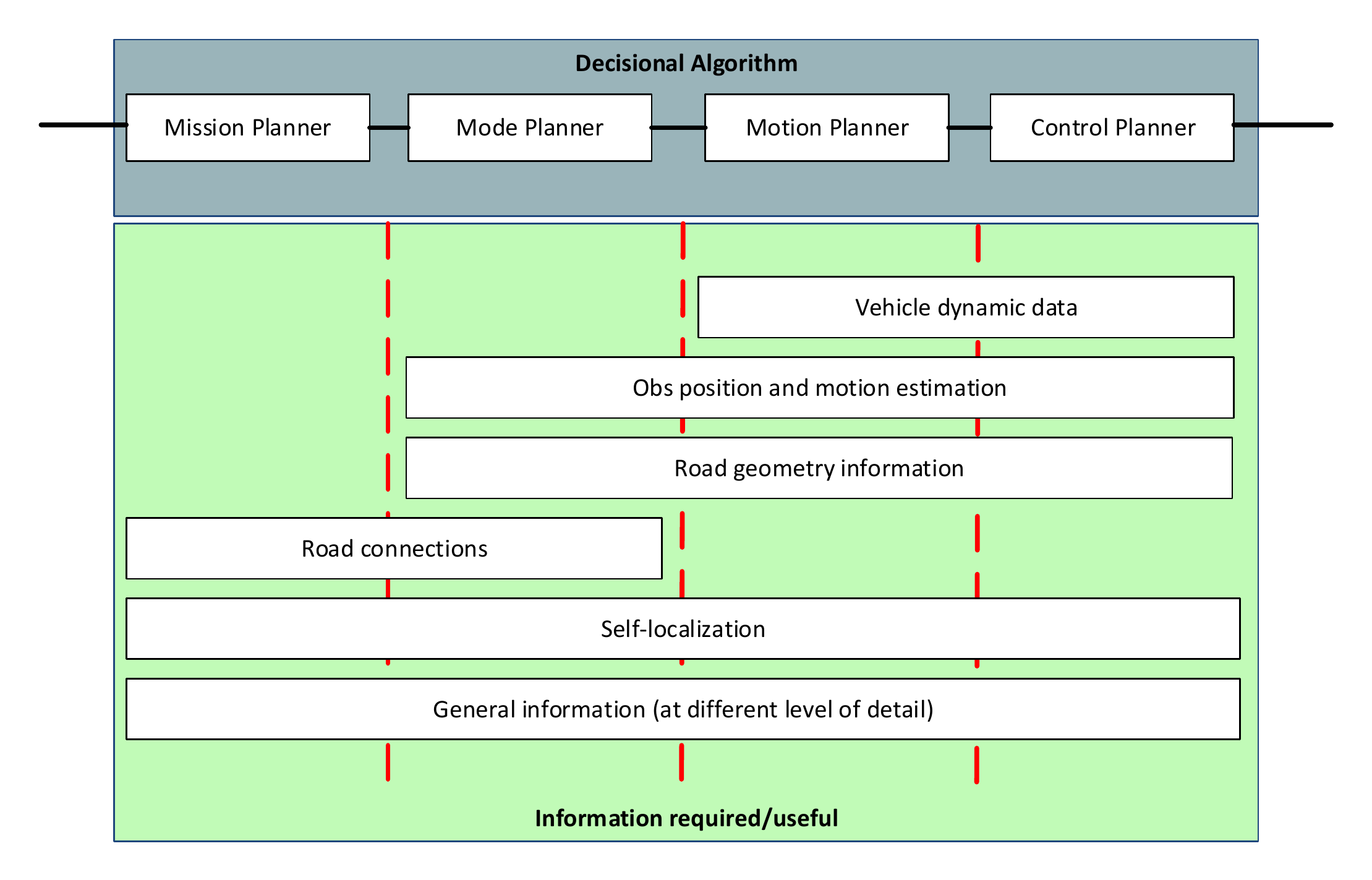}
	\caption{Decisional Algorithm}
	\label{fig:decisional_scheme}
\end{figure}

"Decisional Algorithm", as suggested, incorporates the complex decisional process of human driving. As proposed by previous works \cite{survey_frazzoli,scheme_survey}, it is possible to split the entire process into 4 sub-tasks in a hierarchical scheme as shown in fig.\ref{fig:decisional_scheme}.

The main novelty contribution of this research work is represented by the design of a decisional algorithm capable of performing trajectory planning and vehicle control tasks ensuring:
\begin{itemize}
	\item a real-time feasible implementation;
	\item a sufficient robustness of the solution calculated;
	\item the capability of correctly handle with multiple and moving obstacles in urban-like scenarios.
\end{itemize}

These requirements can be translated into specifics that can be evaluated and verified.

In particular real-time implementation means to define a minimum working frequency for the controller that allows to dynamically react fast enough to unexpected (dangerous) situation that can occur. This limit is set around 20 Hz, that correspond to a reaction time of $0.05s$: this is limited mainly by considering the working frequency of camera sensors (usually around 30 Hz to which should be added the delay due to image processing).
Robustness is intended mainly as the capability of the controller to always generate a feasible result (even if sub-optimal), to be capable of correctly operate although moderate modeling errors or external disturbances and finally to prevent to get stuck in local minima.
Moreover the algorithm is required to correctly compute vehicle control inputs that generate a safe and collision free trajectory in presence of several multiple and moving obstacle. 
It's finally required to operate in a urban-like environment. It is characterized by the presence of sharp curves, narrow lines and medium-low driving speed required. 

In detail, fig. \ref{fig:decisional_scheme} represents the different information required at every decisional tasks: in this work, Mission planner is not further investigated as well as Mode planning (a specific reference trajectory as output of these two tasks is considered known).

"Vehicle dynamic data" (model parameters), "Self-localization", "Road geometry information" and "Obstacle's position and motion estimation" are considered known and with a sufficient accuracy level than the designed algorithms considers these values as deterministic.

\section{Path-planner scheme}\label{sec:PP}

The trajectory planning \& control problem investigated can be defined as shown in \ref{eq:OCP}.

	\begin{equation}\label{eq:OCP}
	\begin{aligned}
		\min_{\mathbf{u}(t)} \quad&L_{tot}({\mathbf{x}}(t),{\mathbf{u}}(t))\\ 
		\text{s.t.}\ &\dot{\mathbf{x}}(t)=f(\mathbf{x}(t),\mathbf{u}(t))\\ 
		&\mathbf{x}(t_0)=\mathbf{x_0}
	\end{aligned}
	\end{equation}
where $L_{tot}$ represents the cost function of the problem including also soft constraints formulation, while $f(\mathbf{x}(t),\mathbf{u}(t))$ represents the numerical model of the vehicle: in the following, optimization constraint problem is described in all its parts. In detail road and vehicle models, constraints and cost function formulations are reported.

\subsection{Map model}\label{section1}
"Road geometry information" are assumed as known data: at every time step the designed algorithm can rely on road structure's knowledge of an area equal or greater that the spatial horizon considered by the optimization problem. Road information are defined as function of the curvilinear abscissa of centerline of the road. Geometrical data and additional data can be distinguished.
Additional data include: number of lanes; left and right width respect to centerline of the lane ($b_{yh}$,$b_{yl}$); and other information that can be spatially located and defined as function of curvilinear abscissa (i.e. traffic lights or traffic signs position, intersections, speed limits,\dots). Geometrical data include: global coordinates of centerline of the road ($X$,$Y$), heading angle of the road ($\theta_c$) and it derivatives ($\theta'_{cs}$).
In order to: 
\begin{itemize}
	\item reduce as much as possible the amount of data required to fully describe the geometrical structure of the road;
	\item minimize the computational effort required to perform a self-localization of the vehicle from curvilinear reference system (electronic map) to global coordinate;
	\item minimize the computational effort required to evaluate "Geometrical data" from curvilinear coordinate.
\end{itemize}
The mathematical structure of the road is defined as a Cubic Hermite Spline (CHS) as reported in \cite{zhang16}.

\begin{figure}[htp]\centering
	\includegraphics[width=1\columnwidth]{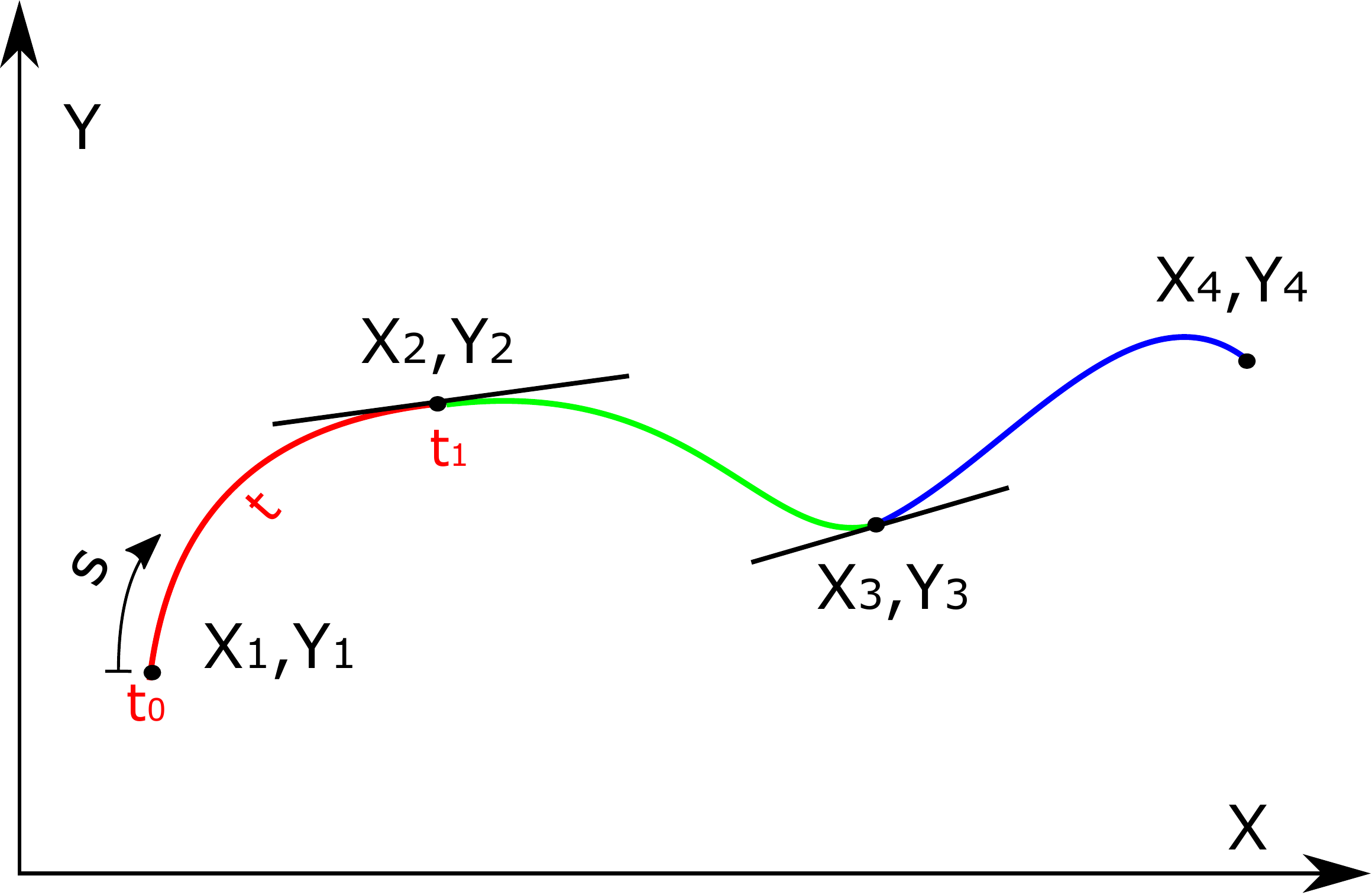}
	\caption{Cubic Hermit Spline example}
	\label{fig:CHS}
\end{figure}

CHS is a continuous function defined as a piecewise, where each sub-function is defined as a cubic polynomial in Hermite form \eqref{eq:CHS_1}. 
\begin{equation}\label{eq:CHS_1}
\begin{aligned}
p(t) =& H_{00}(t)p_0 + H_{10}(t)d_0 + H_{01}(t)p_1 + H_{11}(t)d_1 \quad \\
\text{for}&\quad [s_k,s_{k+1}]\\
\text{where:} &\quad t\in [0,1]\qquad t=\frac{s-s_k}{s_{k+1}-s_k}\\
& p_0 = p(t_0) \qquad d_0 = \frac{\partial p }{\partial t}(t_0) \\
& p_1 = p(t_1) \qquad d_1 = \frac{\partial p }{\partial t}(t_1) \\
& H_{00} = 2t^3-3t^2+1\qquad H_{10} = t^3-2t^2+t\\
& H_{01} = -2t^3+3t^2\qquad  H_{11} = t^3-t^2\\
\end{aligned}
\end{equation}
where $p(t)$ are respectively $X$,$Y$ global coordinates of the centerline of the road and $s$ is the curvilinear coordinate (or operatively the traveled distance on the path calculated from an arbitrary initial point) as shown in fig.\ref{fig:CHS}. In order to define each polynomial parameters, a set of boundary conditions (values and derivatives) in $s_k$ positions must be imposed. In that way is possible to define $X(s)$, $Y(s)$ as continuous functions  and with at least second order continuous derivatives.
The decisional algorithm designed requires also absolute heading angle of the road and its first derivative respect to $s$ (curvilinear coordinate). Considering CHS road representation, these information can be numerically defined by means of geometrical formulas in \eqref{eq:CHS_2}:
\begin{equation}\label{eq:CHS_2}
\begin{aligned}
\theta_c(s)= &\arctan(\frac{Y'_s(s)}{X'_s(s)})\\
\theta'_{cs}(s)= & \frac{X'_sY''_s-Y'_sX''_s}{(X'^2_s+Y'^2_s)^{3/2}}
\end{aligned}
\end{equation}
where the terms $X'_s,Y'_s,X''_s,Y''_s$ are first and second derivatives respect to $s$ of $X(s)$ and $Y(s)$. 

\subsection{Vehicle Description And Modeling}
\begin{figure}[htp]\centering
	\includegraphics[width=1\columnwidth]{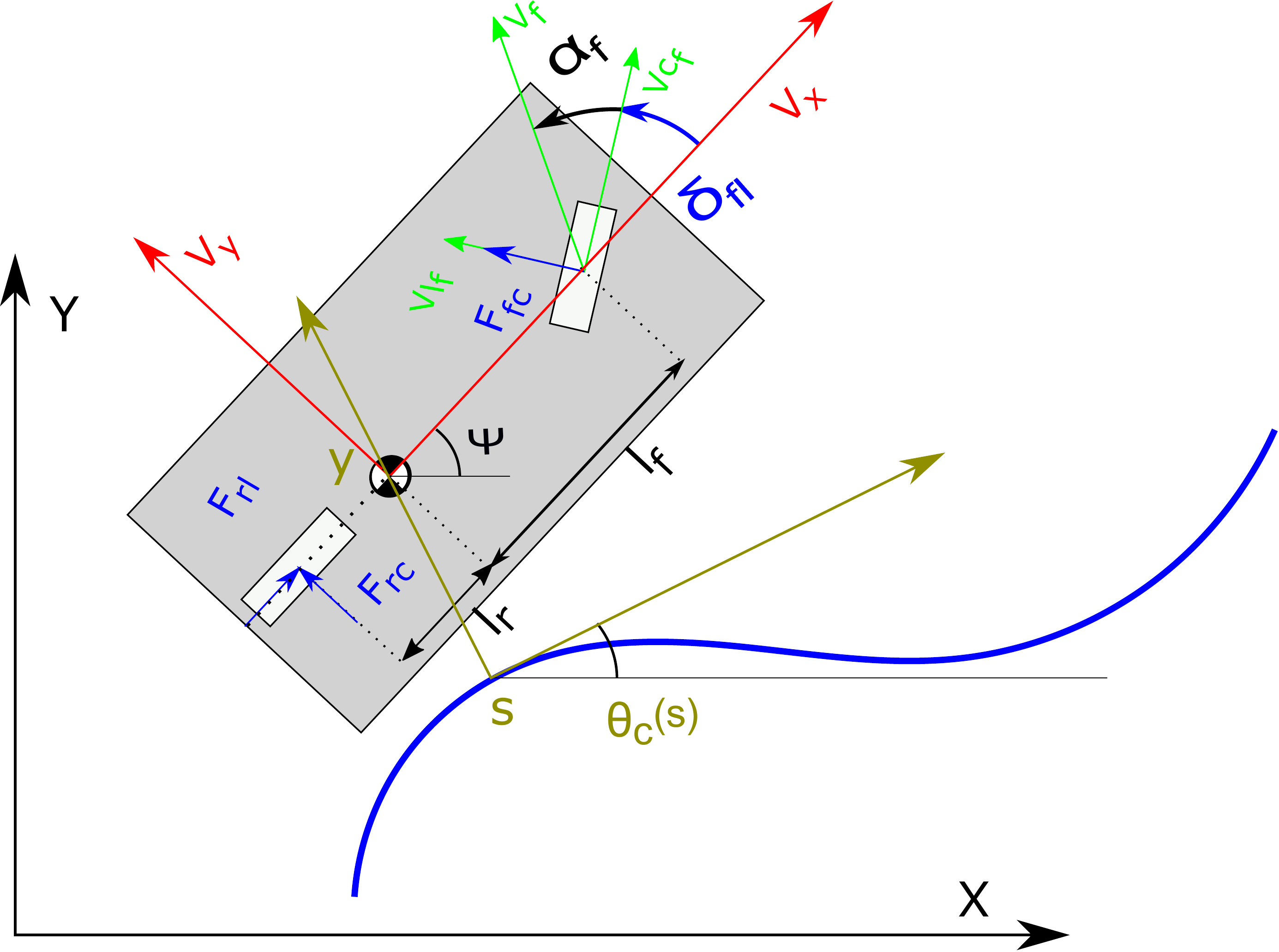}
	\caption{Vehicle model reference system}
	\label{fig:GA_model}
\end{figure}

In fig. \ref{fig:GA_model} a schematization of vehicle model and relative reference systems used is shown. 
State function describing system dynamics is composed by eight state variables ($\mathbf{x}(t)\in\mathbb{R}^{8}$) and two input variables ($\mathbf{u}(t)\in\mathbb{R}^{2}$). State function can be compactly written as $\dot{\mathbf{x}}(t)= f^{GA}(\mathbf{x}(t),\mathbf{u}(t))$. 
In detail, as reported in \eqref{eq:GA_model2}, state $\mathbf{x}(t)$ is composed by curvilinear coordinate and lateral displacement respect to road centerline ($s$,$y$), absolute heading angle $\psi$, longitudinal and lateral speed in the body frame ($V_x$,$V_y$), yaw rate ($\omega$), front wheel steering angle ($\delta$) and finally torque applied to the rear wheels ($\tau$).
Input variables $\mathbf{u}(t)$ are the derivatives of wheel steering angle  and torque applied, defined for clarity as $u_1 = \dot{\delta}$,$u_2 = \dot{\tau}$ respectively.

\begin{equation}\label{eq:GA_model2}
\begin{aligned}
\mathbf{x}(t) & =[s,y,\psi,V_x,V_y,\omega,\delta,\tau]^T\\
\mathbf{u}(t) & =[u_1,u_2]^T\\
\end{aligned}
\end{equation} 

Dynamic vehicle's formulation shown in \eqref{eq:GA_model} is based on a single-track simplification. 

\begin{equation}\label{eq:GA_model}
\left\{
\begin{aligned}
\dot{s} = & [\cos (\theta_c)(V_x\cos (\psi) - V_y\sin (\psi)) + \sin (\theta_c)(V_x\sin (\psi) + \\
& + V_y\cos (\psi))]/(1 - \theta'_{cs}y)\\
\dot{y} = & [-\sin (\theta_c)(V_x\cos (\psi) - V_y\sin (\psi)) + \cos (\theta_c)(V_x\sin (\psi) + \\
& + V_y\cos (\psi))]\\
\dot{\psi}= & \omega\\
\dot{V}_x = & \omega V_y + \frac{2F_{rl}}{M} - \frac{2F_{fc}\sin (\delta)}{M} - \frac{F_{aero}}{M}\\
\dot{V}_y = & -	\omega V_x  + \frac{2F_{rc}}{M} + \frac{2F_{fc}\cos (\delta)}{M}\\
\dot{\omega} = & 2/J(-F_{rc}l_r + F_{fc}l_f\cos(\delta))\\
\dot{\delta}= & u_1\\
\dot{\tau}= & u_2\\
\end{aligned}
\right.
\end{equation}
Where $\theta_c$ and $\theta'_{cs}$ are absolute heading angle of road's centerline and the derivative respect to curvilinear coordinate $s$ (these values are considered known), $M$ and $J$ are total mass and inertia respect to z axis of the vehicle,  $l_f$ and $l_r$  are relative distances between front and rear axles  respect to CoG.
in \eqref{eq:GA_model}:
\begin{itemize}
	\item  $F_{aero}$ represents aerodynamic resistance defined as function proportional to ${V}_x^2$;
	\item $F_{rl}$ corresponds to longitudinal force applied to the vehicle assuming that only rear wheels are actuated. In detail, vehicle powertrain is neglected and longitudinal force is simplified as $F_{rl} = {\tau}/{R_r}$, where $R_r$ represents wheel radius;
	\item $F_{fc}$ and $F_{rc}$ are lateral forces generated by tire-road interaction modeled by Pacejka's non-linear numerical model  ("Magic formula"), as reported in \eqref{eq:GA_model3}:
\end{itemize}

\begin{equation}\label{eq:GA_model3}
F_{\ast c}(\alpha_\ast) = -D\sin(C\arctan(B\alpha_\ast + E(\arctan(B\alpha_\ast)-B\alpha_\ast)))
\end{equation}
where $B,C,D,E$ are Pacejka model parameters and $\alpha_\ast$ is slip angle of the wheel.
Front and rear slip angles are computed by \eqref{eq:GA_model4}, where the additional term $\epsilon$ is defined to numerically prevent indetermination due to stationary vehicle condition ($V_x = 0$) and it's mathematically implemented as $\epsilon = \epsilon_0e^{-V_x}$.
\begin{equation}\label{eq:GA_model4}
\begin{aligned}
(\alpha_f) =& (V_y + l_f\omega)/(V_x + \epsilon) -\delta\\
(\alpha_r) =& (V_y - l_r\omega)/(V_x + \epsilon)
\end{aligned}
\end{equation}

Vertical load $F_z$ is considered as a constant value defined by an equal distribution of vehicle's mass on each wheel as $F_z = Mg/4$.

\subsection{Constraints}\label{sec:GA_Con}
 The optimization problem presents a set of inequality constraints \eqref{eq:GA_const1}.
 In detail, constraints are implemented on state variables, inputs and relative position respect to obstacles (defined by mean of a special formulation deeply described in next section).
 \begin{equation}\label{eq:GA_const1}
 \begin{aligned}
 &\text{State constrants}\\
 &b_{yl} \le y \le b_{yh}\\
 &b_{\delta l} \le \delta \le b_{\delta h}\\
 &b_{\tau l} \le \tau  \le b_{\tau h}\\
 &\text{Inputs constrants}\\
 &b_{\dot{\delta} l} \le \dot{\delta} \le b_{\dot{\delta} h}\\
 &b_{\dot{\tau} l} \le \dot{\tau}  \le b_{\dot{\tau} h}\\
 &\text{Obstacles constrants}\\
 &dist_{rel_i} \ge 0\\
 \end{aligned}
 \end{equation}
 Each variable is constrained within a feasible range bounded by a set of lower and upper values (i.e on $y$ variable $b_{yl}$ and $b_{yh}$ correspond to lower and upper maximum lateral width of the lane respect to centerline).
 Constraints considered in the proposed formulation act on lateral displacement of the vehicle (maintain the lane), on maximum steering angle allowed, maximum accelerating and braking torque, on steering angle and torque ratio allowed and finally on relative distance of the vehicle respect to obstacles present on the lane.
 The set of constraints presented are evaluated for each discretized time-step which compose the time horizon of the optimization problem.
 
 Constraints are mathematically formulated as equality equations by means of slack variables as in \eqref{eq:GA_const2}: each state and input variable is constrained by two equality equations (and slack variables), while for each obstacle constraint equation is defined by mean of the relative distance between vehicle and i-th obstacle. All constraints are evaluated as soft constraints in the cost function as shown in next section.
 \begin{equation}\label{eq:GA_const2}
 \begin{aligned}
 &Z_1 = y - b_{yl}\\
 &Z_2 = -y + b_{yh}\\
 &\dots\\
 &Z_{Obs_i} = dist_{rel_i}\\
 \end{aligned}
 \end{equation}
 
 \subsection{Obstacles' constraints}\label{sec:GA_Constr}
 
 While state and inputs constraints are directly defined on $\mathbf{x},\mathbf{u}$ of the problem as reported in \eqref{eq:GA_const1}, obstacles constraints require the evaluation of relative distance between vehicle and obstacles ($dist_{rel_i}$).
 That value depends by self-localization of the vehicle ($\mathbf{x}$)  and obstacle's position ($\mathbf{x_{Obs}} = [s_{Obs},y_{Obs},\theta_{Obs}]^T$) on the curvilinear reference system evaluated for each time step considered $dist_{rel_i}(\mathbf{x},\mathbf{u},\mathbf{x_{Obs}})$.
 \begin{figure}[htp]\centering
 	\includegraphics[width=0.8\columnwidth]{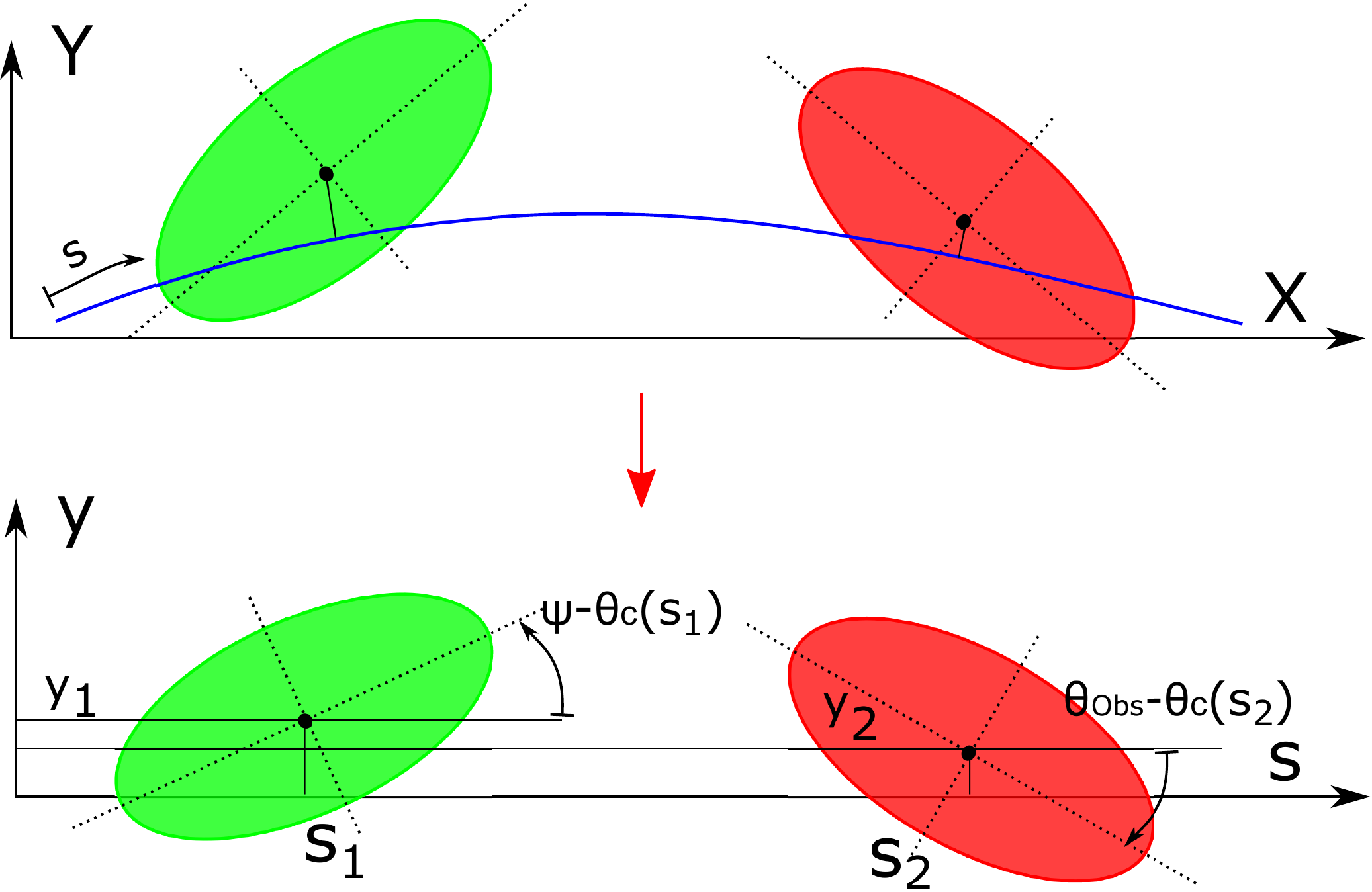}
 	\caption[Vehicle-Obstacle reference system transformation]{Vehicle-Obstacle projection from global to curvilinear reference system.}
 	\label{fig:OBS}
 \end{figure} 
 Considering both measurements ($\mathbf{x}$, $\mathbf{x_{Obs}}$) as deterministically known, obstacles and vehicle are simplified as ellipses, where vehicle dimensions are known, while obstacles' dimensions are assumed as available (i.e. estimated). In order to reduce computational effort, relative distance calculation is performed in curvilinear reference system as shown in fig. \ref{fig:OBS}.
 
\begin{figure}[htp]\centering
 	\includegraphics[width=0.9\columnwidth]{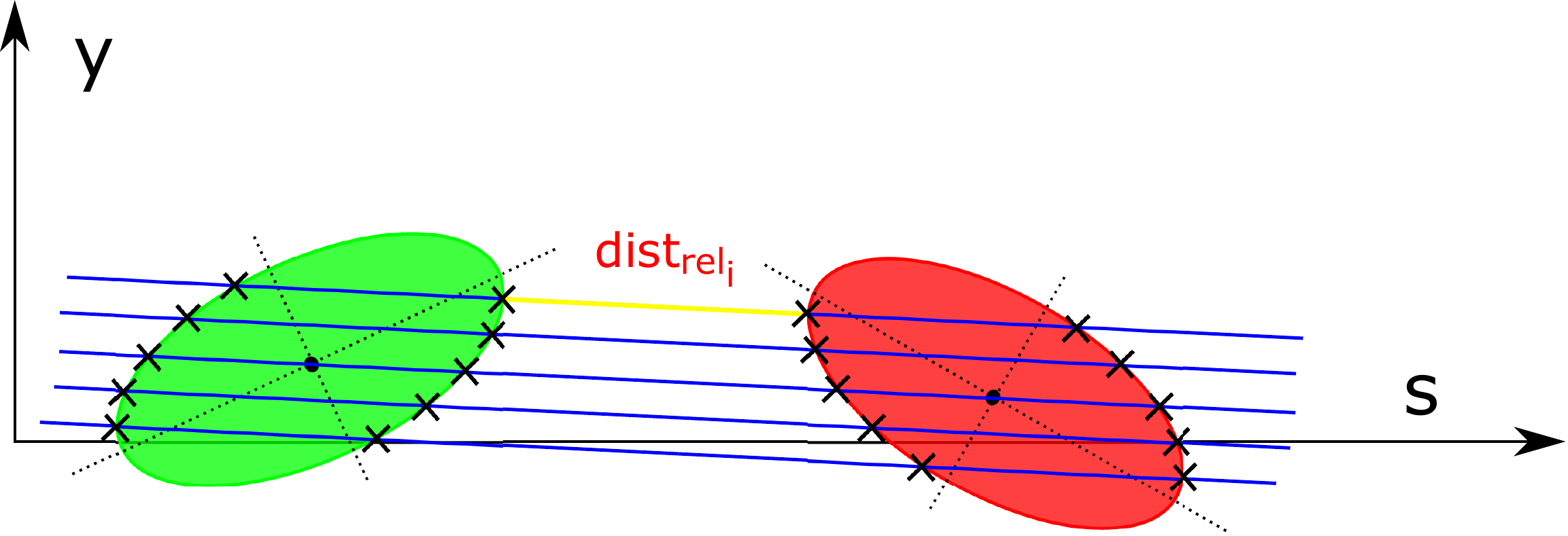}
 	\caption[Relative distance calculation procedure]{Relative distance calculation procedure: blue lines represent straight parallel lines generated, while yellow line is the shortest distance chosen as $dist_{rel_i}$.}
 	\label{fig:OBS2}
 \end{figure}

 In order to evaluate the relative distance $dist_{rel_i}$, an approximate strategy is implemented and shown in fig \ref{fig:OBS2}:
 \begin{itemize}
 	\item a straight line connecting centers of the two ellipses is defined;
 	\item a finite family of parallel lines (i.e. five in fig \ref{fig:OBS2}) is generated. Line relative distances are selected in order to pass through $1/3$ and $2/3$ of semi-minor axis (or semi-major depending on the relative position of the two bodies);
 	\item intersections between ellipses and lines are computed;
 	\item relative distances between intersection nodes are calculated;
 	\item shortest distance is chosen as $dist_{rel_i}$ value.
 \end{itemize} 
 
 For every time step obstacles' position is required. This information is measured just at first time-step and estimated in the rest of predictive horizon. Assuming that actual obstacles' speed is provided, obstacles' position in time-steps subsequent the first is simply obtained by assuming them moving at constant speed. This assumption is made considering that working frequency of the algorithm is high enough (so also obstacles' measures) and predictive horizon of the logic is limited (less than $3$ seconds).
 
\subsection{Cost function}
 The total cost function of the optimization problem ($L_{tot}$) is composed by several terms as presented in \eqref{eq:GA_cost1}:
 \begin{equation}\label{eq:GA_cost1}
 \begin{aligned}
 L_{tot} =& L_N + \frac{1}{2}\sum_{k=0}^{N-1} \{L_{state_k}(\mathbf{W},\mathbf{x})\}dt +\\ &+\frac{1}{2}\sum_{k=0}^{N-1} \{L_{U_k}(\mathbf{S},\mathbf{u})\}dt + \sum_{k=0}^{N-1} \{L_{cnstr_k}(\mathbf{P},\mathbf{Z})\}dt
 \end{aligned}
 \end{equation}
 In detail $L_N$ is the terminal cost evaluated in the last discretized time-step of optimization horizon; $L_{state_k}$ are the cost terms associated to state variables deviation respect to the reference and their relative importance is defined by means of a set of weights represented by matrix  $\mathbf{W}$; $L_{U_k}$ are cost terms associated to optimization variables magnitude and weighted by means of matrix  $\mathbf{S}$; finally $L_{cnstr_k}$ are penalty functions associated to the constraints dependent by weighting coefficients $\mathbf{P}$ and slack variables $\mathbf{Z}$ (in the form defined in the previous section). 
 
 More in detail, $L_{state_k}$ (shown in \eqref{eq:GA_cost2})  considers several aspects of the maneuver: 
 \begin{itemize}
 	\item deviation respect to reference maneuver with terms: $y$ lateral displacement respect to centerline and  $V_x-V_{x\_ref}$ speed error respect to reference;
 	\item comfort-like aspect of the performed maneuver with terms: $V_y$ lateral speed in body frame reference system, $\omega$ yaw rate, $\delta$ steering angle and $\tau$ applied torque.
 \end{itemize}
 \begin{equation}\label{eq:GA_cost2}
 \begin{aligned}
 L_{state_k} = &w_2y^2 + w_4(V_x-V_{x\_ref})^2 + w_5V_y^2 + w_6\omega^2 + \\ 
 &+w_7\delta^2 + w_8\tau^2
 \end{aligned}
 \end{equation} 
 
 $L_{U_k}$ (shown in \eqref{eq:GA_cost3})  considers the terms related to inputs (the derivatives of $\delta$ and $\tau$) that directly affects comfort of the maneuver (their value affect the jerk of the maneuver).
 \begin{equation}\label{eq:GA_cost3}
 L_{U_k} = s_1{u_1}^2 + s_2{u_2}^2 
 \end{equation}
 All terms present in $L_{state_k}$ and $L_{U_k}$ are defined by means of weighted quadratic terms. 
 
 $L_{cnstr_k}$ terms are penalty function terms associated to the constraints. As already stated, all constraints are evaluated as soft constraints. In detail in \eqref{eq:GA_cost4}, all equality constraints associated to state and inputs variables, are introduced in the cost function by means of exponential functions. This mathematical formulation is implemented in order to obtain negligible values when constraints are satisfied, while high values (and with a high gradient) when close to constraint border or breaking it.
 \begin{equation}\label{eq:GA_cost4}
 \begin{aligned}
 L_{cnstr_k} = & e^{(1-p_{y l}Z_1)} + e^{(1-p_{y h}Z_2)} + e^{(1-p_{\delta l}Z_3)} + e^{(1-p_{\delta h}Z_4)}+\\
 &+ e^{(1-p_{\tau l}Z_5)} + e^{(1-p_{\tau h}Z_6)} + e^{(1-p_{u_1 l}Z_7)} + e^{(1-p_{u_1 h}Z_8)} +\\
 &+ e^{(1-p_{u_2 l}Z_9)} + e^{(1-p_{u_2 h}Z_10)} + L_{Obs}
 \end{aligned}
 \end{equation}
 In order to obtain the desired behavior, parameters of weighting matrix $\mathbf{P}$ are accordingly tuned.
 In \eqref{eq:GA_cost5} penalty term related to obstacle is fully described. It differs from other penalty functions because it's defined as a reciprocal function. 
 \begin{equation}\label{eq:GA_cost5}
 L_{Obs} = \sum_{i=1}^{nObs} \frac{p_{Obs}}{Z_{Obs_i} + \epsilon_{Obs}}
 \end{equation}
 
 In order to avoid division by zero when relative distance is equal to zero ($Z_{Obs_i}=0$), an additional parameter (close to zero) ($\epsilon_{Obs}$) is considered. Moreover this term is composed by contribution of relative distance evaluation respect to all obstacles present on the road. This mathematical formulation guarantees negligible values when far from zero relative distance, while extremely high values when close to zero (that is minimum value possible for $Z_{Obs_i}$).
 
 Finally the terminal cost $L_N$ is presented in \eqref{eq:GA_cost6}.
 \begin{equation}\label{eq:GA_cost6}
 L_N = \frac{1}{2}\{L_{state_N}(\mathbf{W},\mathbf{x})\}dt +\{L_{cnstr_N}(\mathbf{P},\mathbf{Z})\}dt
 \end{equation}

\section{Genetic Algorithm implementation}\label{sec:GA}
\begin{figure}[htp]\centering
	\includegraphics[width=1\columnwidth]{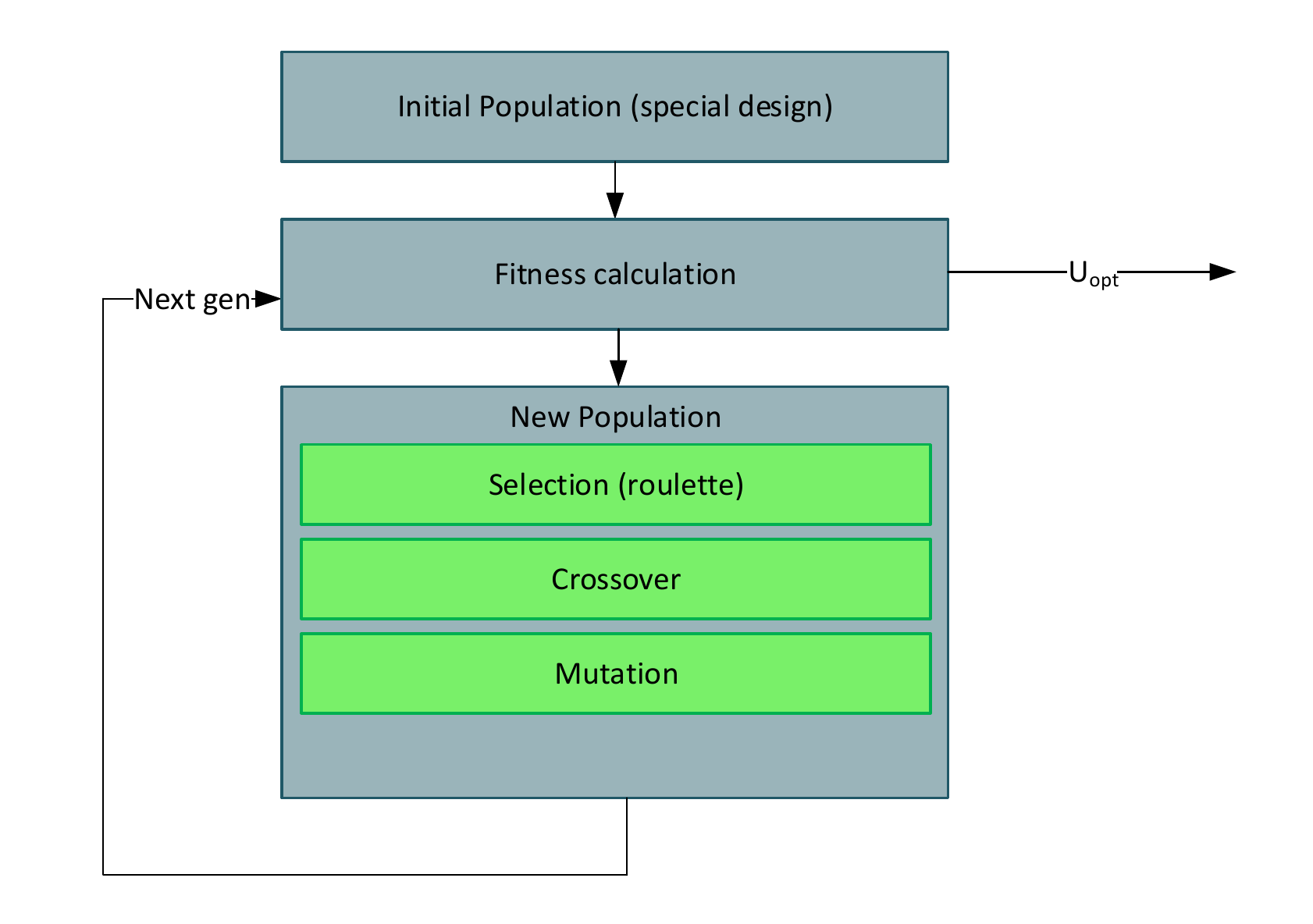}
	\caption{General GA architecture}
	\label{fig:GA_1}
\end{figure}

Genetic Algorithm (GA) can be classified as an adaptive heuristic search technique loosely inspired by the evolutionary ideas of natural selection and genetics. They can be considered a smart exploitation of a random search applied on optimization problems \cite{EngOptimMeta,PSO_NatureInsp,metaheur}. Although a randomized mechanism, GAs uses "evolutionary information" to direct the search to the optimal solution contained within the search space. GAs family of search algorithms are designed to simulate the natural evolutionary process through steps similar to the principles introduced by the scientist Charles Darwin. 
GA presents several advantages over classical optimization techniques as to be applicable to discontinuous, non-linear and not differentiable cost functions (its implementation does not involve differential calculation or gradient evaluation).
A classical implementation of GA technique (high order population and unbounded number of generations stopped when a condition has been reached) does not fit with fast calculation requirements, but on the other hand the iterative optimization through an evolutionary process allow to stop the process when the maximum allowed calculation time is reached obtaining a feasible suboptimal solution.
Starting from a general GA architecture as shown in the scheme of fig. \ref{fig:GA_1} and based on classical GA schemes reoprted in \cite{bookeur}, the specific formulation of each task is described in detail in the following.

\subsection{Mathematical formulation of the problem}

Each candidate solution $\mathbf{U}$ of the optimization problem is defined as a sequence of $N$ discretized values of input variables ($u_1$,$u_2$) as in \eqref{eq:GA_prob1}.
\begin{equation}\label{eq:GA_prob1}
\mathbf{U}^T = 
\begin{Bmatrix}
u_1^1 & u_1^2 & \dots & u_1^N \\
u_2^1 & u_2^2 & \dots & u_2^N \\
\end{Bmatrix}
\end{equation}
GA implementation designed, is based on the assumption of a linear trend of each actuation variable over the horizon (time continuity of actuation commands). This allows to define the candidate solution as in \eqref{eq:GA_prob2}.
\begin{equation}\label{eq:GA_prob2}
\begin{aligned}
\mathbf{U}^T = 
\begin{Bmatrix}
a_1*i+ b_1\\
a_2*i+ b_2 \\
\end{Bmatrix} \quad i=1,\dots,N
\end{aligned}
\end{equation}
This formulation reduces the independent variables that can freely vary from $2N$ to $4$ with a dramatic reduction of computational complexity of the optimization problem. Moreover a continuous input sequence (linear dependency) force the dynamics to have a continuous jerk trend (with benefits in term of comfort). On the other hand, this assumption intrinsically limit the system to sub-optimal solution. However high working frequency of the controller and a reduced length of optimization horizon limits the sub-optimality.
Another implementing decision is the continuous value encoding of the variables (instead of the most commonly used binary encoding). This choice is done in order to eliminate the discretization effect due to the "number of bits" of encoding process and to include zero value within the search space.

The iterative process is stopped after a fixed number of "Generations" $N_{gen}$: the result of the process are the coefficients of the "best" solution candidate. The sub-optimal control sequence resulting is computed as \eqref{eq:GA_prob9}:
\begin{equation}\label{eq:GA_prob9}
\begin{aligned}
\mathbf{\tilde{U}}^T = 
\begin{Bmatrix}
\tilde{a_1}*i+ \tilde{b_1}\\
\tilde{a_2}*i+ \tilde{b_2} \\
\end{Bmatrix} \quad i=0,\dots,N-1
\end{aligned}
\end{equation}

\subsection{Initial population}\label{sec:GA_init_popul}

\begin{figure}[htp]\centering
	\includegraphics[width=0.7\columnwidth]{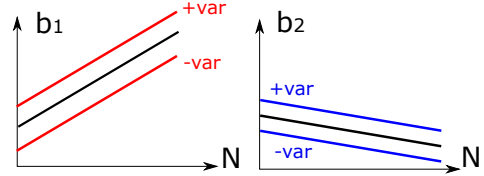}
	\caption{Variational solutions}
	\label{fig:GA_2}
\end{figure}

The design of the initial population is an important aspect that highly affects the possibility to find a feasible solution and the convergence speed to find sub-optimal solutions close to the best possible.
The implementation proposed is composed by three sub-sequential phases:

\paragraph{Previous solution} 
the first candidate solution is defined as equal to the sub-optimal solution calculated at the previous iteration of the algorithm;
\paragraph{Variational solutions}
starting from "previous solution" defined as in \eqref{eq:GA_prob2}, a set of "variational solutions" is obtained considering small variations for the zero-order terms ($b_1,b_2$). The variations considered are $[-var,0,+var]$, where $var$ is relative variation respect to the nominal value (fig. \ref{fig:GA_2}). 
"Variational solutions" are obtained considering all possible combinations. 
In the specific implementation $3^2 = 9$;
\paragraph{Random candidate solutions}
The rest of candidate solutions composing the initial population are calculated by randomly define set of values inside a feasible search space defined by boundaries as in \eqref{eq:GA_prob3}.
\begin{equation}\label{eq:GA_prob3}
\begin{aligned}
b_\bullet &= k*lim_{u_\bullet} \qquad \bullet\in(1,2) \qquad k\in[1;1]\\
a_\bullet &= k*lim_{u_\bullet}- b_\bullet/(Ndt) \quad\qquad k\in[1;1]\\
\end{aligned}
\end{equation}
where $lim_{u_\bullet}$ is a limit value for input variables: it can be coincident or more conservative respect to constraint value defined in section \ref{sec:GA_Con} and $Ndt$ corresponds to total time horizon of the optimization.

\subsection{Fitness calculation}

Fitness calculation is defined by mean of cost function evaluation over the horizon for each solution candidate in actual population. In detail, fitness calculation is defined as the reciprocal of cost function $L_{tot}$ because minimization of cost function correspond to maximization of fitness performance (as typically is implemented in GA).
\begin{equation}\label{eq:GA_prob4}
fitness_i = \frac{1}{L_{tot}(U_i)}
\end{equation}
Fitness calculation allows to rank each candidate solution in term of "adaptation" and to define the global fitness of the population $fitness_{tot}$ in \eqref{eq:GA_prob5}.
\begin{equation}\label{eq:GA_prob5}
fitness_{tot} = \sum_{i=1}^{N_{pop}}fitness_i
\end{equation}

\subsection{Selection}

Selection process has the main purpose to select the best candidates among population to be the "genetic material" used to generate a new population.
The selection is usually driven by fitness-ranking evaluation by means of a specific mathematical method. There exist many different approaches in literature as i.e. Boltzman selection, rank selection, steady state selection, \dots \ref{bookeur}

\begin{figure}[htp]\centering
	\includegraphics[width=1\columnwidth]{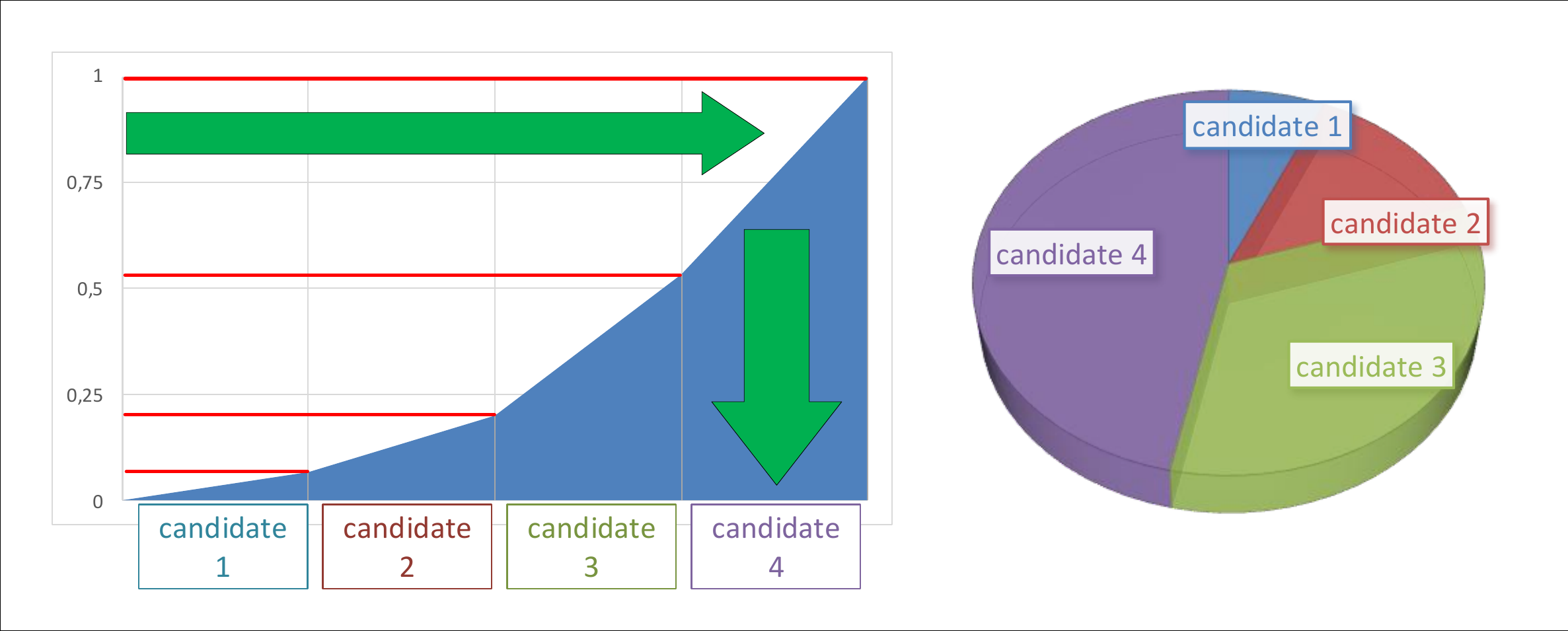}
	\caption{Roulette Wheel Selection example}
	\label{fig:GA_3}
\end{figure}

The proposed implementation of the decisional algorithm designed is based on "Roulette Wheel Selection".
The mathematical principle can be easily explained by roulette wheel analogy: all solution candidates of population have a "slice" of the wheel, whose width is depending by its relative fitness respect to the total as in \eqref{eq:GA_prob6}: 
\begin{equation}\label{eq:GA_prob6}
width = \frac{fitness_i}{fitness_{tot}}
\end{equation}
Defined a cumulative function of slice widths, using a randomly selected value $\in[0;1]$, it's possible to associate a solution candidate from actual population. The process, described by means of an example in fig. \ref{fig:GA_3}, is iterated until $N_{pop}$ candidates are selected.

\subsection{Crossover}

Crossover (and also mutation) generates a completely new population considering as "parents" the solution candidates chosen by "Selection" process. This approach eliminates good solution candidates from actual population (that could have been better than the new ones just generated). This drawback can slow down convergence to expected sub-optimal solution.
This reason crossover algorithm designed also implements an "elitism" mechanism.
Elitism technique consists in maintaining the best candidate solution (or a few of them) into the new population.
In detail crossover algorithm implemented is composed by three phases (similar to the ones presented in "Initial population" sect. \ref{sec:GA_init_popul}).

\paragraph{Elitism} 
the first candidate solution of the new population is defined as equal to the best candidate solution present in actual population (best fitness value).
\paragraph{Variational solutions}
as already described in sect. \ref{sec:GA_init_popul}, a set of "variational solutions" around the best candidate solution is computed considering the $9$ possible combinations.
\paragraph{Continuous crossover}
the rest of new population is calculated by a linear combination of parents solution candidates. The linear combination is computed by a weighted mean where the weight coefficient is randomly defined as $\in[0;1]$. The process presented in \eqref{eq:GA_prob7} for a generic variable $x$ is repeated for each optimization variable ($a_1,b_1,a_2,b_2$).
\begin{equation}\label{eq:GA_prob7}
\begin{aligned}
x_{new_1} &= (\alpha_x)x_{old_1} + (1-\alpha_x)x_{old_2}\\
x_{new_2} &= (1-\alpha_x)x_{old_1} + (\alpha_x)x_{old_2}\\
\end{aligned}
\end{equation}
where $x_{new_1},x_{new_2}$ are the new variable values, $x_{old_1},x_{old_2}$ are "parents" values of the variable and finally $\alpha_x$ is the weight coefficient associated.

\subsection{Mutation}

Mutation is implemented as a random change of an optimization variable after crossover task is completed. This task allows to better explore the entire search space, but at the same time a too high mutation rate may reduce the convergence process to optimality in favor of a completely random search.
As in Crossover task, elitism is implemented: first candidate solution of the generated population from crossover task remains equal to the best candidate solution present in actual population.
All other solution candidate optimization variables are evaluated and modified by \eqref{eq:GA_prob8}.
\begin{equation}\label{eq:GA_prob8}
\begin{aligned}
&if(\beta_x\ge \beta_{th})\\
&\qquad x_{new} = (1+\alpha_{mut})x_{old}\\
&else\\
&\qquad x_{new} =x_{old}\\
&end\\
\end{aligned}
\end{equation}
Mutation is applied if a random parameter $\beta_x$ is greater than a threshold $\beta_{th}$.
Mutation is computed by means of a multiplication (random) factor $\alpha_{mut}$ selected within a limited mutation range.

After mutation stage a new population is finally generated and represents initial population of a new "generation".

\section{Simulation results}\label{sec:res}

\subsection{Testing Area and Map generation}\label{sec:res_map}

\begin{figure}[htp]\centering
	\includegraphics[width=0.95\columnwidth]{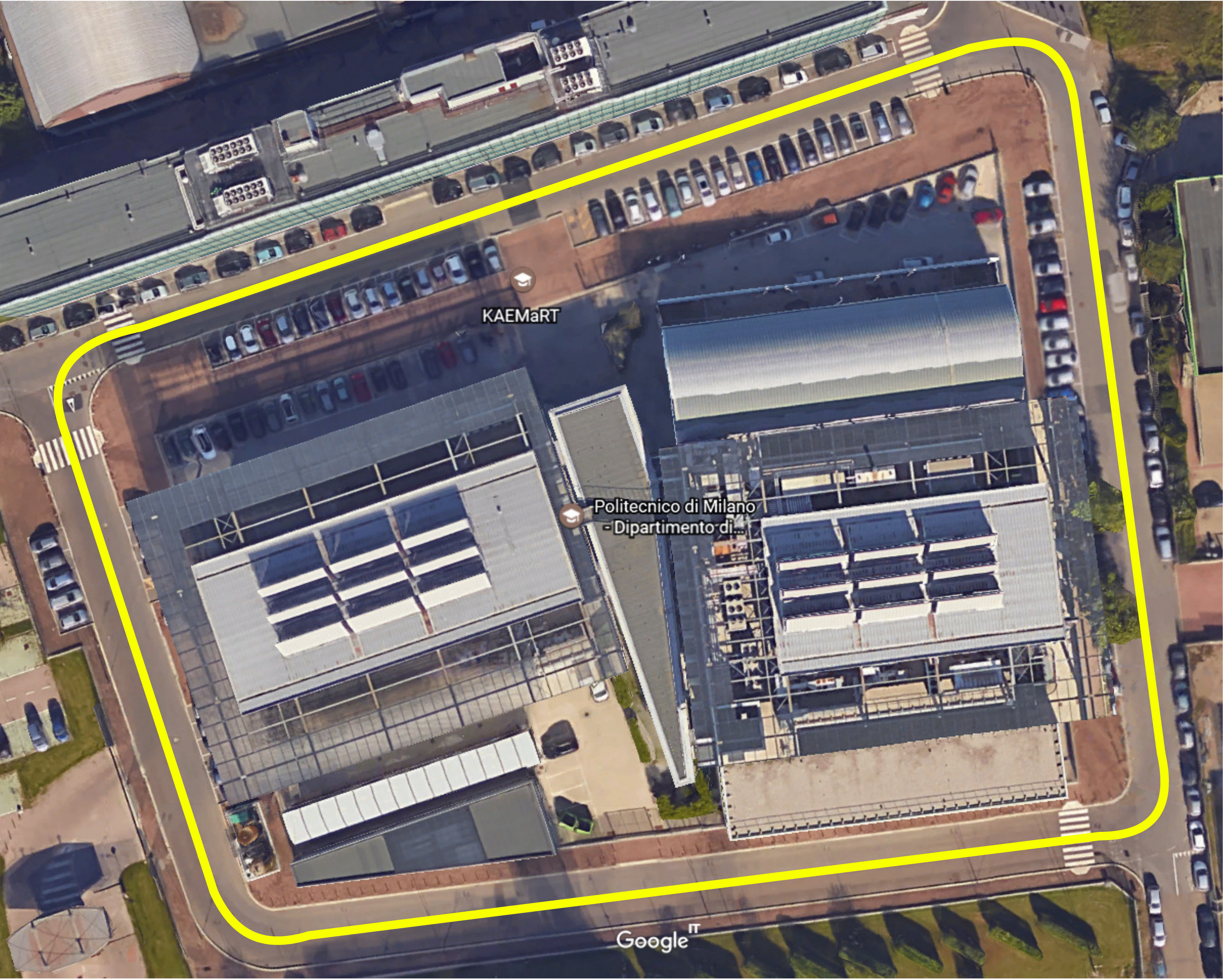}
	\caption{Mechanical Engineering Department area}
	\label{fig:res_1}
\end{figure}

In order to simulate the proposed algorithm in a realistic scenario that can represent typical urban traffic conditions, experimental campaigns have been conducted.
The testing area selected, is the area surrounding Mechanical Engineering Department of Politecnico di Milano as shown in fig.\ref{fig:res_1}.

\begin{figure}[htp]\centering
	\includegraphics[width=0.95\columnwidth]{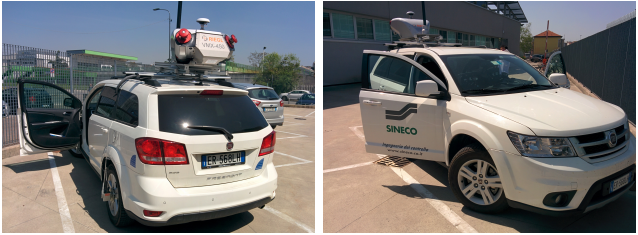}
	\caption[SINECO S.p.A experimental vehicle]{SINECO S.p.A experimental vehicle equipped with RIEGL VMX-450 measurement system}
	\label{fig:res_2}
\end{figure}

Thanks to SINECO S.p.A support \cite{sineco}, an experimental campaign was conducted in order to obtain accurate measurements of GPS track of road centerline. The experimental vehicle (shown in fig. \ref{fig:res_2}) is equipped with RIEGL VMX-450 measurement system. This system is composed by laser scanners, IMU/GNSS equipment, and cameras. IMU/GNSS system is capable of position measurements at 200 Hz, with an extremely high accuracy (measurement uncertainty about $2cm$). Moreover laser scanners and cameras can provide accurate 3D point clouds and images of the surrounding area but are not of interest for th experiment.
The experimental campaign was performed by manually driving the instrumented vehicle in the center of the road in order to obtain a GPS sequence of the centerline of the street.
The mapped area presents four straights and four sharp curves. The total distance of an entire lap is around $450m$ and total GPS points collected (representing road centerline) are about $12000$.

In order to define a CHS structure suitable for the decisional algorithm starting from GPS row data collected, lane geometry approximation is performed as presented in \cite{zhang16} by the author. 
The points considered in CHS interpolation are selected by resampling GPS data collected considering a fixed traveled distance between subsequent points. $ds$ is set to $4m$ that generates a reduction of overall considered points to just $114$.

As suggested in \cite{zhang16}, positions are used as locations of control points of spline in Hermite form. Tangent vectors ($\mathbf{v_i}$), instead, are computed by means of a minimization of error between generated curve and reference measured trajectory defined (considering Hausdorff distance calculation as accuracy index) as in eq.\ref{eq:res_1}. 
\begin{equation}\label{eq:res_1}
\min_{v_i,\dots,v_N} \sum_{k=1}^{N_{\hat{p}}} \lVert f_{CHS}(u_k|\mathbf{S_0})-\mathbf{\hat{p_k}}\rVert^2
\end{equation}
where $f_{CHS}(u_k|\mathbf{S_0})$ represents the CHS curve parameterized by $\mathbf{S_0}$ $\{(u_i \mathbf{\bar{p_i}} \mathbf{v_i})|i\in [1,N_{\bar{p}}] \}$ where $\mathbf{p_i}$ are control points locations, $u_i$ are knots of CHS, $\mathbf{v_i}$ are the tangent vectors to be determined and finally $N_{\hat{p}} $ and $N_{\bar{p}}$ are total collected and resampled points respectively.
\begin{figure}[htp]\centering
	\includegraphics[width=0.95\columnwidth]{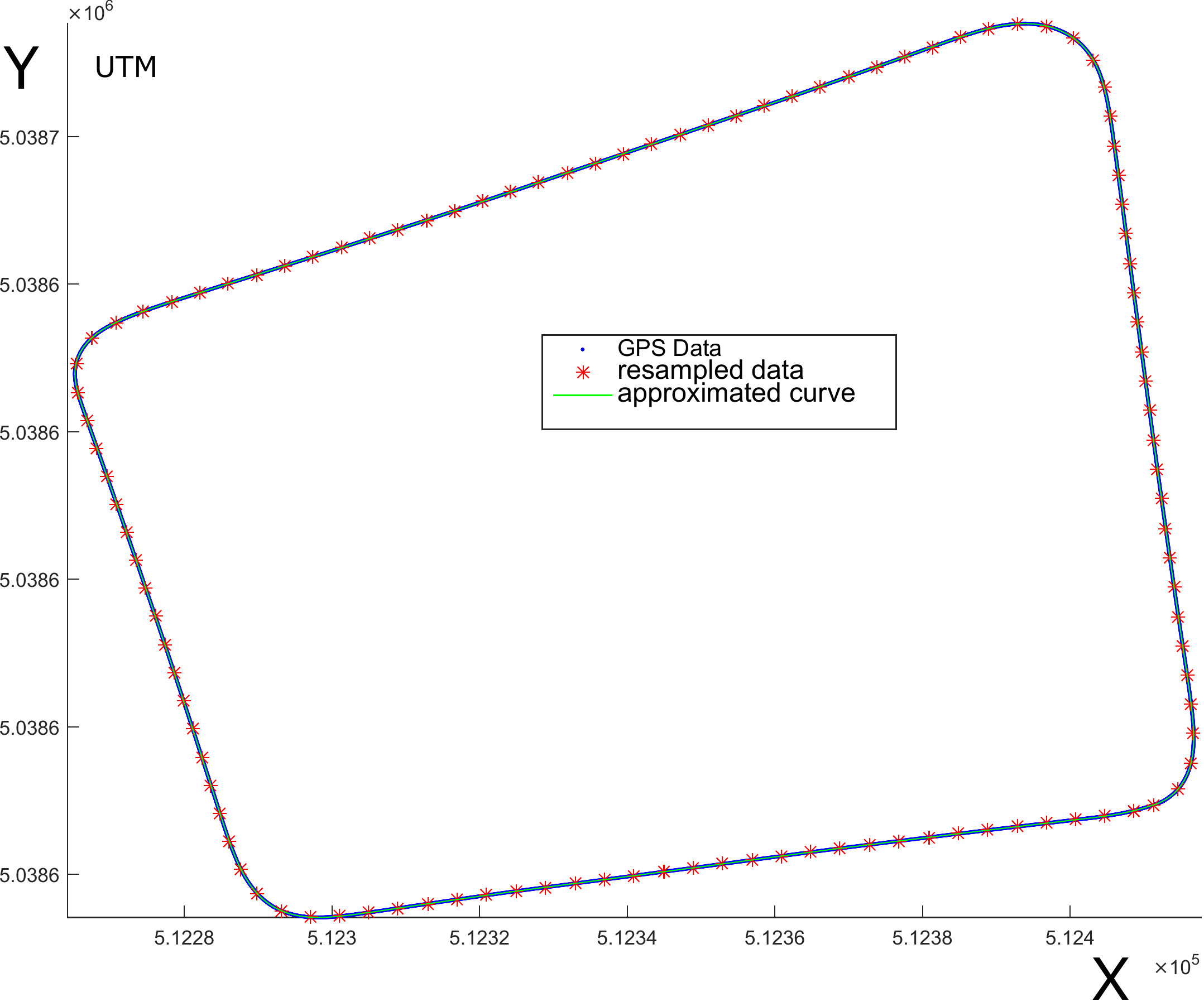}
	\caption[Simplified Mechanical Engineering Department map]{Simplified Mechanical Engineering Department map, where blue dots are raw GPS data, red dots are resampled GPS points and green line is the approximate resulting curve}
	\label{fig:res_3}
\end{figure}

Applying the simplification method described, it's possible to obtain the map shown in fig. \ref{fig:res_3}. In detail blue and red dots represent raw and resampled GPS data, while the green line is the resulting CHS curve. In particular the simplified CHS curve presents a Hausdorff maximum distance of about $0.03m$ respect to the original raw GPS collected data (considered as the "correct" centerline curve). 
The mapped area consists in a two lanes urban road (used as single lane in out tests). Lane width is fixed in $2.75m$, so total road width is around $5.5m$. All simulations proposed in the following section consider this value as constant road width inside problem constraints and vehicle reference is set to be the center of the road itself.

\subsection{Numerical simulations}\label{sec:ref_numsim}

\begin{figure}[htp]\centering
	\includegraphics[width=0.95\columnwidth]{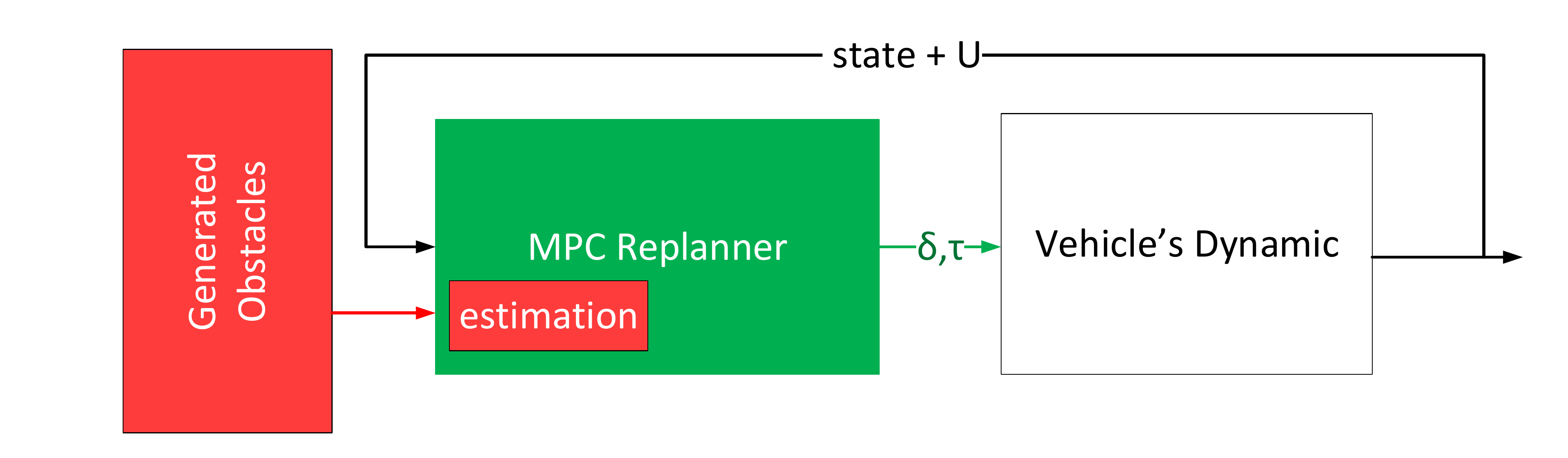}
	\caption{Simulation system architecture}
	\label{fig:res_4}
\end{figure}

The numerical simulations presented in the following are obtained by the closed-loop system architecture defined in fig.\ref{fig:res_4}. A numerical function ("Generated Obstacles") responsible of generation of sensors-like data regarding obstacles (global position $s_{Obs}$,$y_{Obs}$,$\theta_{Obs}$ and speed $V_{sObs}$,$V_{yObs}$ in curvilinear reference frame) is defined. Within the designed replanner, these data are used to estimate obstacles' motion over time horizon of the controller by means of a simple linear extrapolation ("estimation") as proposed in \ref{sec:GA_Constr}. The resulting physical actuation controls ($\delta,\tau$) are set as inputs of "Vehicle's Dynamic". It represents simulated vehicle behavior and it's defined by the same vehicle's model presented in \eqref{eq:GA_model} and used in the decisional algorithm.
"MPC Replanner" (comprehensive of GA Replanner + obstacle's motion estimation) works at $20Hz$ while "Vehicle's Dynamic" is internally evaluated with a discretization time step of $0.01s$.

An optimization time horizon of ($N = 20$) steps with a time discretization of $\Delta t =0.12s$ but state evolution is computed with a discretization of $\Delta t_{state} =0.06s$ is considered.
Vehicle dimensions are assumed as $4m$ (length) and $1.9m$ (width), similar to subcompact car segment. 

The algorithm formulation is maintained fixed (in term of cost function weights and GA parameters) for all the simulation proposed in the following. Thanks to its numerical structure based on a fixed number of iterations, the computational time required is almost fixed. The testing machine considered is a laptop with in Intel-i7 CPU and 4 GB of RAM. The software is developed on MATLAB environment using Microsoft Windows OS.

\begin{figure}[htp]\centering
	\includegraphics[width=0.95\columnwidth]{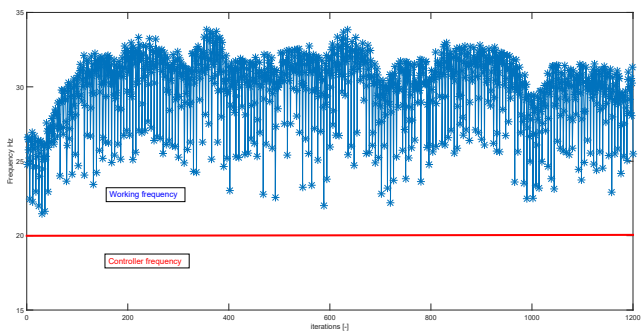}
	\caption[Algorithm frequency]{Algorithm frequency: blue dots represent algorithm estimated maximum working frequency at each time step, while red line is working frequency adopted for the controller}
	\label{fig:res_14}
\end{figure}

Fig. \ref{fig:res_14} shown the algorithm working frequency during a generic simulation: the working frequency due to computational effort required is always higher $20Hz$ confirming that it's possible to apply it in a real-time application. The computational effort is not scenario dependent.
\subsubsection{Complete lap}\label{sec:res_complete}

The first simulations presented consist in a full lap of the testing scenario. 

\begin{figure}[htp]\centering
	\includegraphics[width=0.95\columnwidth]{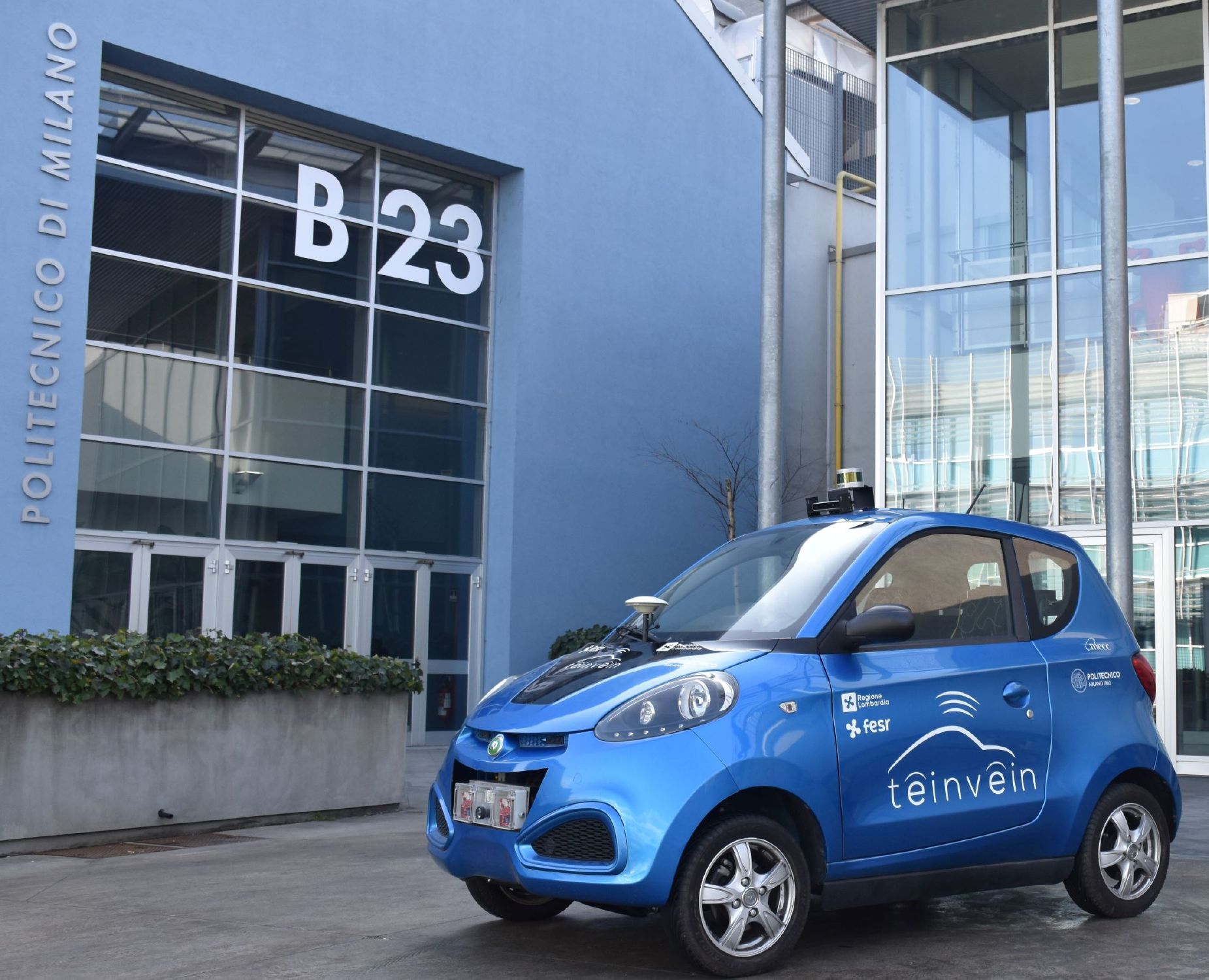}
	\caption[prototype vehicle]{Prototype vehicle under development \cite{teinvein}}
	\label{fig:teinvein}
\end{figure}

The simulations are performed by defining vehicle reference speed equal to $4m/s$ and $8m/s$ respectively. Lower speed ($4m/s$) is proposed in order to compare simulation results to future experimental tests scheduled using a prototype vehicle under development \cite{teinvein}(fig.\ref{fig:teinvein}).
Higher speed is defined as a realistic value for the area selected (urban area among building in a university campus).
\begin{figure}[htp]\centering
	\includegraphics[width=0.95\columnwidth]{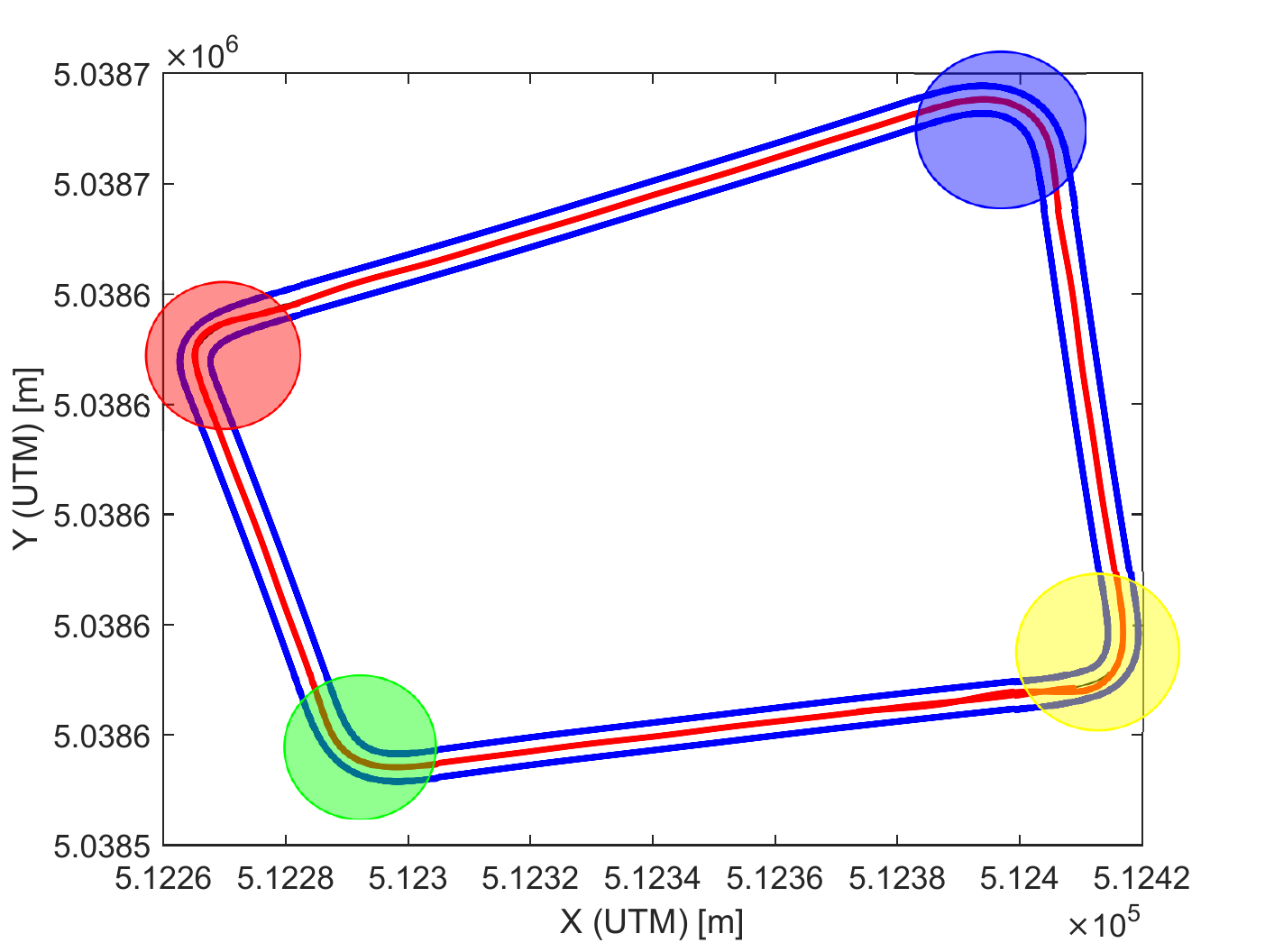}
	\caption[Complete lap trajectory]{Complete lap simulated: where red line represents the computed trajectory,blue lines represent lane limits and colored areas represents turns}
	\label{fig:res_7}
\end{figure}
In fig.\ref{fig:res_7} a top view of the numerical simulation performed at $4m/s$ is shown. Cornering areas are the most challenging part of the scenario and are highlighted by colored circles in order to identify them in the following graphs reported.

\begin{figure}[htp]\centering
	\includegraphics[width=0.95\columnwidth]{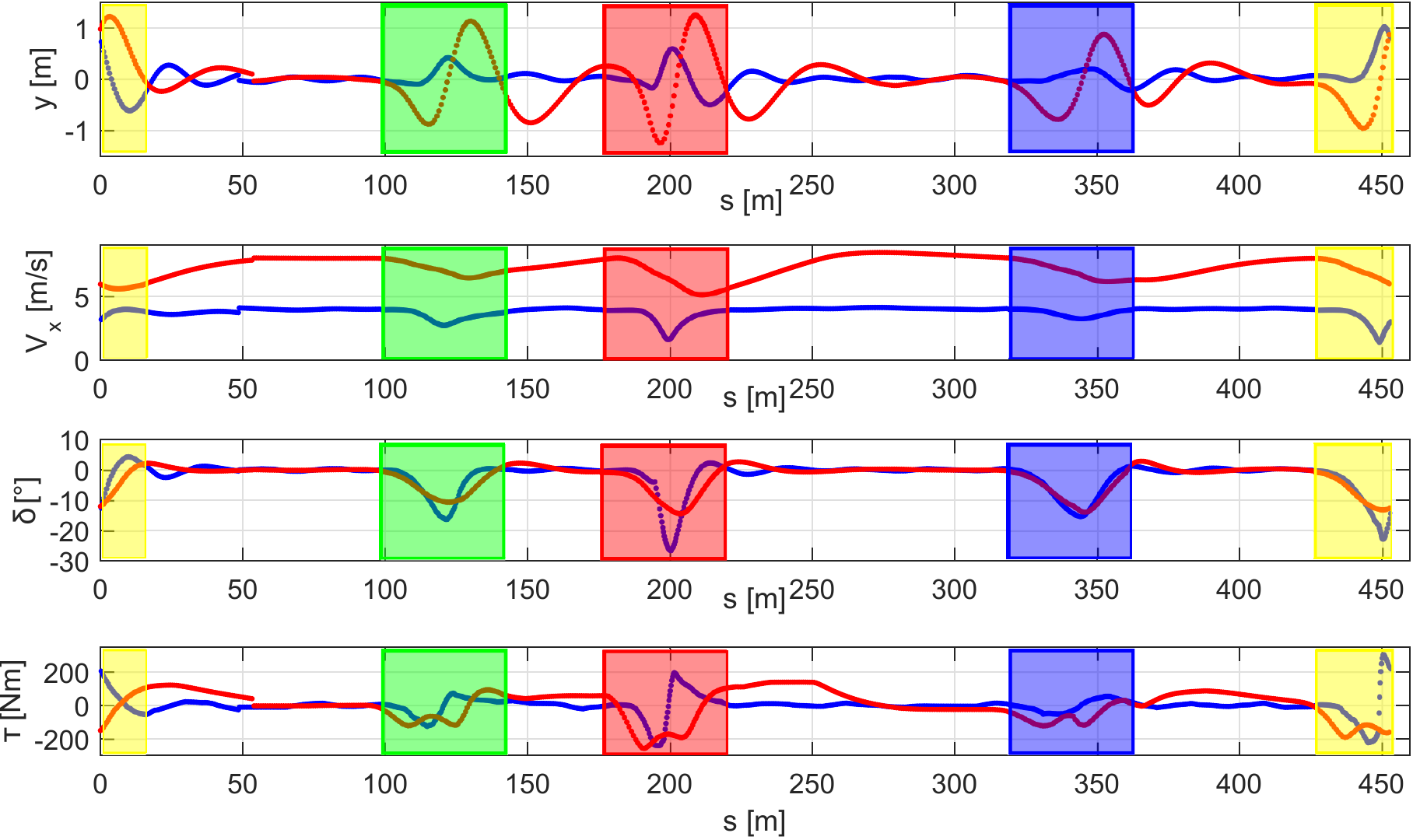}
	\caption[Complete lap simulation results]{Complete lap simulation results, where red and blue lines represent time evolution of relevant parameters for high and low reference speed respectively}
	\label{fig:res_8}
\end{figure}
\begin{figure}[htp]
	\centering
	\subfloat[][global reference\label{fig:res_8_a}]
	{\includegraphics[width=0.95\columnwidth]{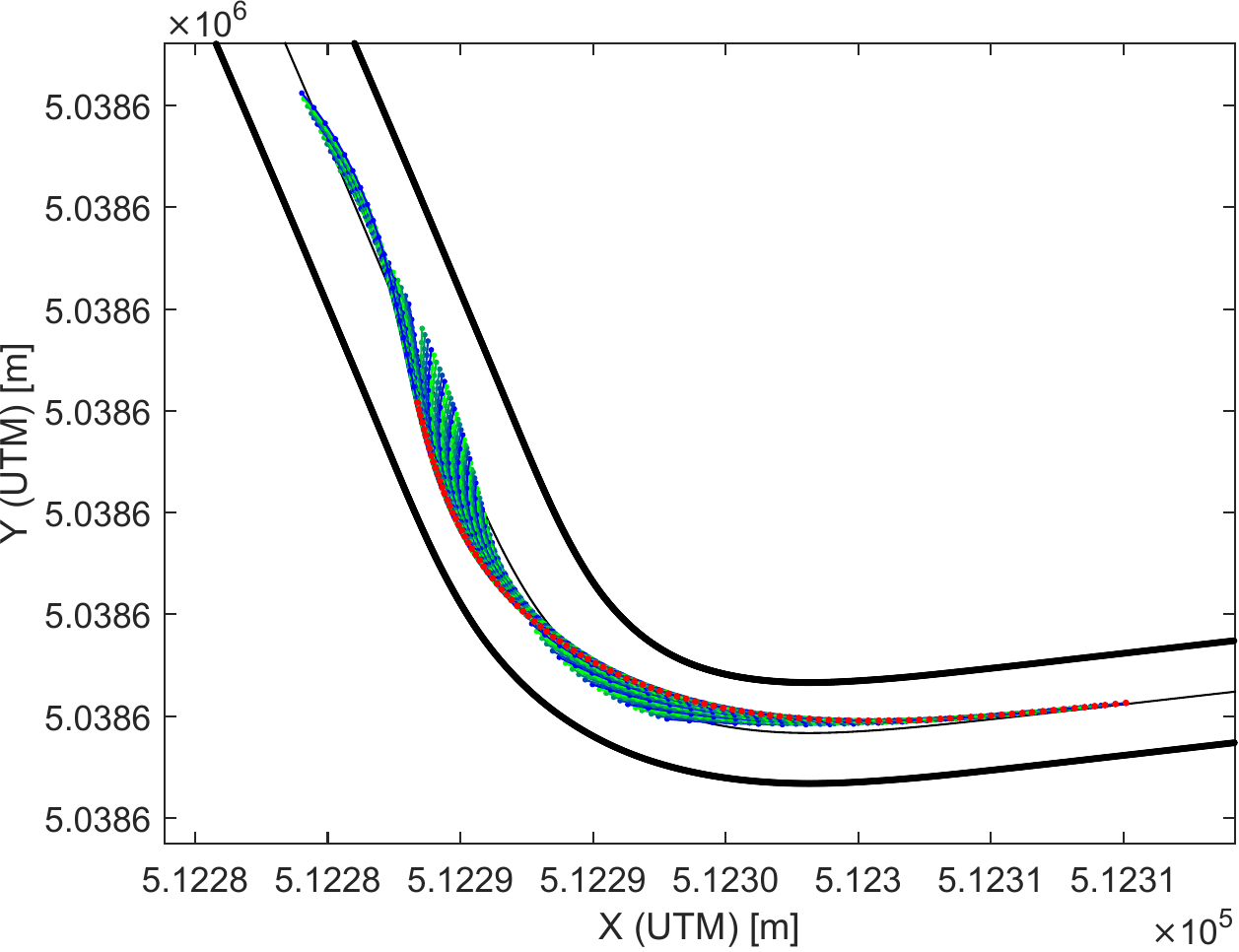}} \\
	\subfloat[][curvilinear reference\label{fig:res_8_b}]
	{\includegraphics[width=0.95\columnwidth]{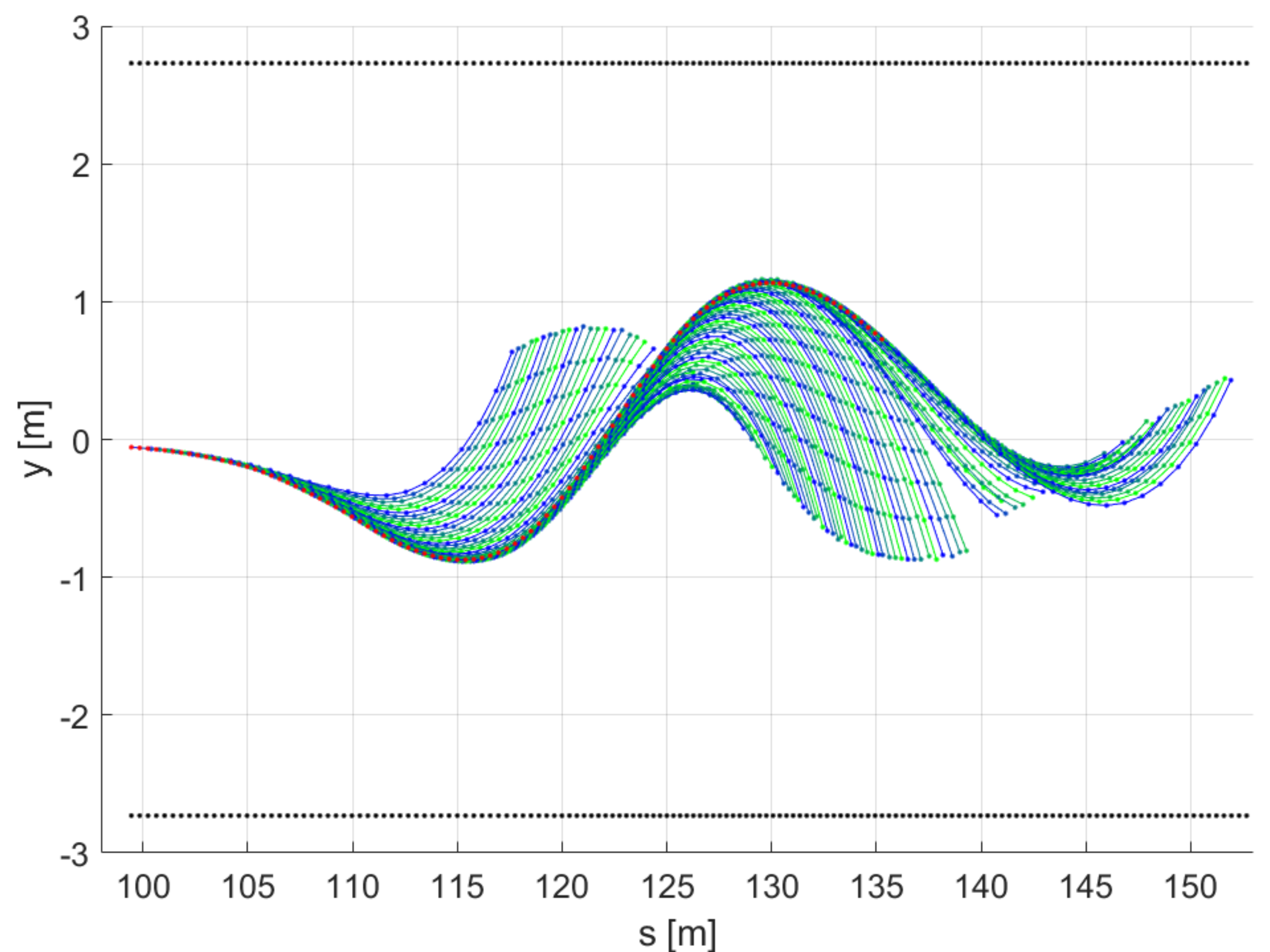}} 
	\caption[Optimized trajectories - turn 1]{Optimized trajectories at turn 1: lines represent optimal trajectories computed at each time-step, while red dots represent the closed loop overall trajectory traveled by the vehicle}
	\label{fig:res_8t}
\end{figure}

In fig. \ref{fig:res_8} evolution of the simulation parameters of interest are reported. In the order, rows represent: lateral displacement respect to road centerline $y$, longitudinal speed in vehicle's reference frame $V_x$ and finally physical actuation commands (steering angle $\delta$ and applied torque $\tau$). In detail blue lines represent slower speed simulation, while red higher speed one. To clearly compare the simulations, all parameters are reported respect to traveled distance $s$ used as x-axis. Moreover turns are highlighted with the same colors used in fig. \ref{fig:res_7}. 
Both cases are correctly handled by the controller: lateral displacement remains below boundary limits imposed to vehicle's CoG ($1.75m$). As expected, high speed maneuver generates an higher lateral displacement due to lateral speed cost term $V_y$ and inputs rate terms $\dot{\delta},\dot{\tau}$. On the other hand an higher speed increases the corresponding spatial predictive horizon of the controller: this is reflected by anticipated maneuvers approaching turns. In detail steering maneuvering anticipate the turn by generating a trajectory that "cuts the corner" maintaining an average higher speed and reducing steering rate.
This effect is also visible from fig.\ref{fig:res_8_a}. 
Both figures in \ref{fig:res_8t} show the turning approach from a time point of view: colored lines represent the optimal trajectories computed by the planner at each time-step, while red dots represent the closed loop overall trajectory traveled by the vehicle. In detail, fig. \ref{fig:res_8_a} is defined in the global coordinate system and shows the "cut the corner" behavior, while fig.\ref{fig:res_8_b} is represented in curvilinear reference as internally used by the controller to compute the optimization. In both figures is possible to notice how at each time-step a new solution of the problem generates a slightly different solution: this is mainly due to the shift of the predictive horizon.

%

\subsubsection{modeling error response}

in order to evaluate control behavior in presence of modeling errors, in this section, tyre to road adhesion coefficient of vehicle model used inside decisional algorithm and the one used in "Vehicle's Dynamic" of the scheme shown in fig.\ref{fig:res_4}  are different on purpose. This mismatch it's a common but dangerous situation in particular during raining or snowing road conditions. The common approach in ADAS algorithms is to define a conservative (or estimated) constant value for it. This can lead to a mismatch between real and expected behavior of the vehicle.

In detail a common and conservative value of friction coefficient for dry asphalt surface defined as $\mu=0.7$ is internally set. Tyre to road adhesion coefficient of simulated vehicle ("vehicle's dynamic") instead is set to values varying in a range defined as $\mu\in[0.2,0.7]$.
The simulations are performed imposing a reference speed equal to $8 m/s$.
In order to clearly present a comparison of the different simulation results, only limited areas of the scenario are shown. 

\begin{figure}[htp]\centering
	\includegraphics[width=0.95\columnwidth]{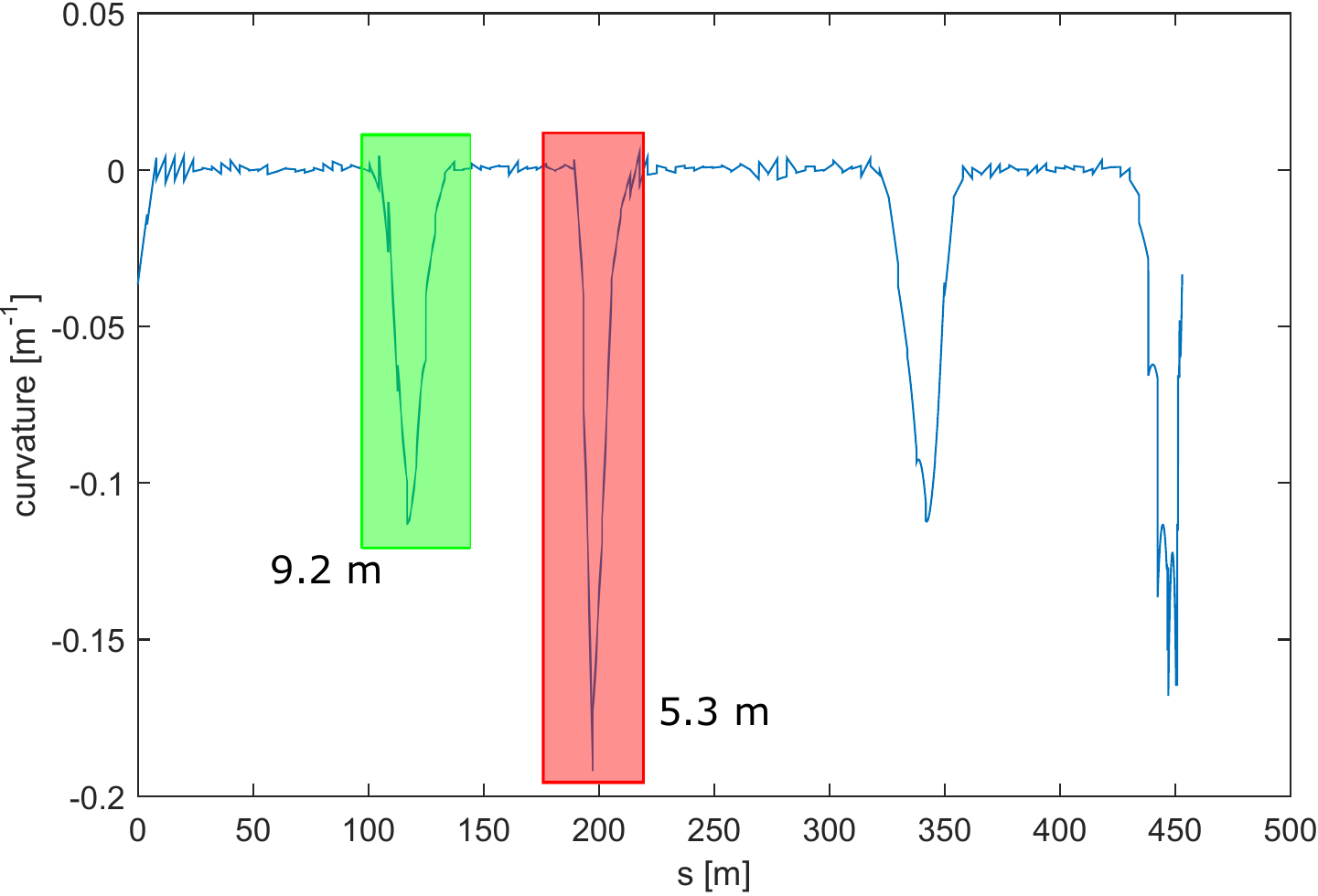}
	\caption{Curvature values of testing scenario's turns}
	\label{fig:res_9}
\end{figure}

First and second turns are chosen as most significant areas considering the graph in fig. \ref{fig:res_9} representing curvature of the track : turn 1 (green) is the smoothest one (curvature radius = $9.2m$) while turn 2 (red) is the sharpest one (curvature radius = $5.3m$).

\begin{figure}
	\centering
	\subfloat[][turn 1\label{fig:res_10}]
	{\includegraphics[width=0.95\columnwidth]{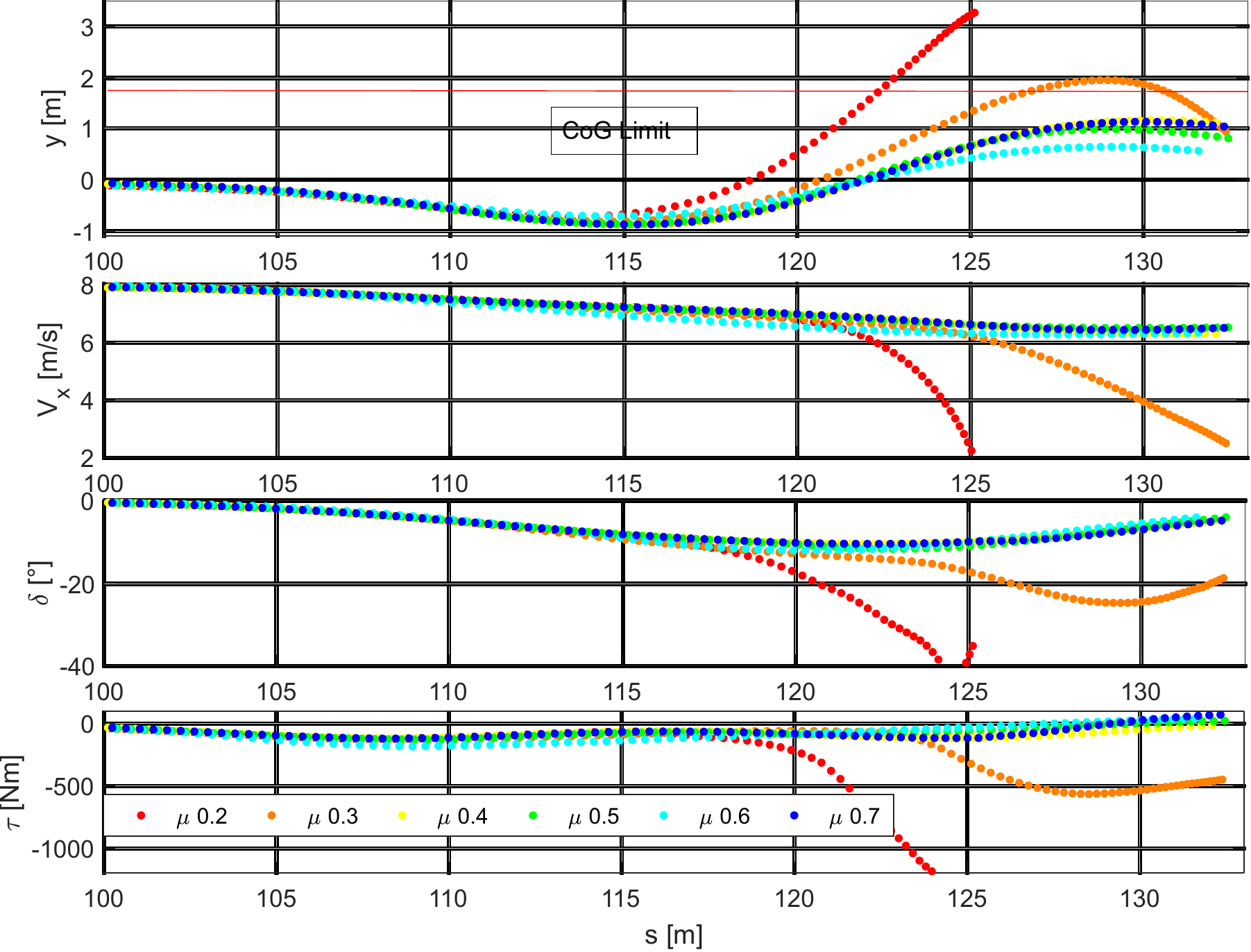}} \\
	\subfloat[][turn 2\label{fig:res_11}]
	{\includegraphics[width=0.95\columnwidth]{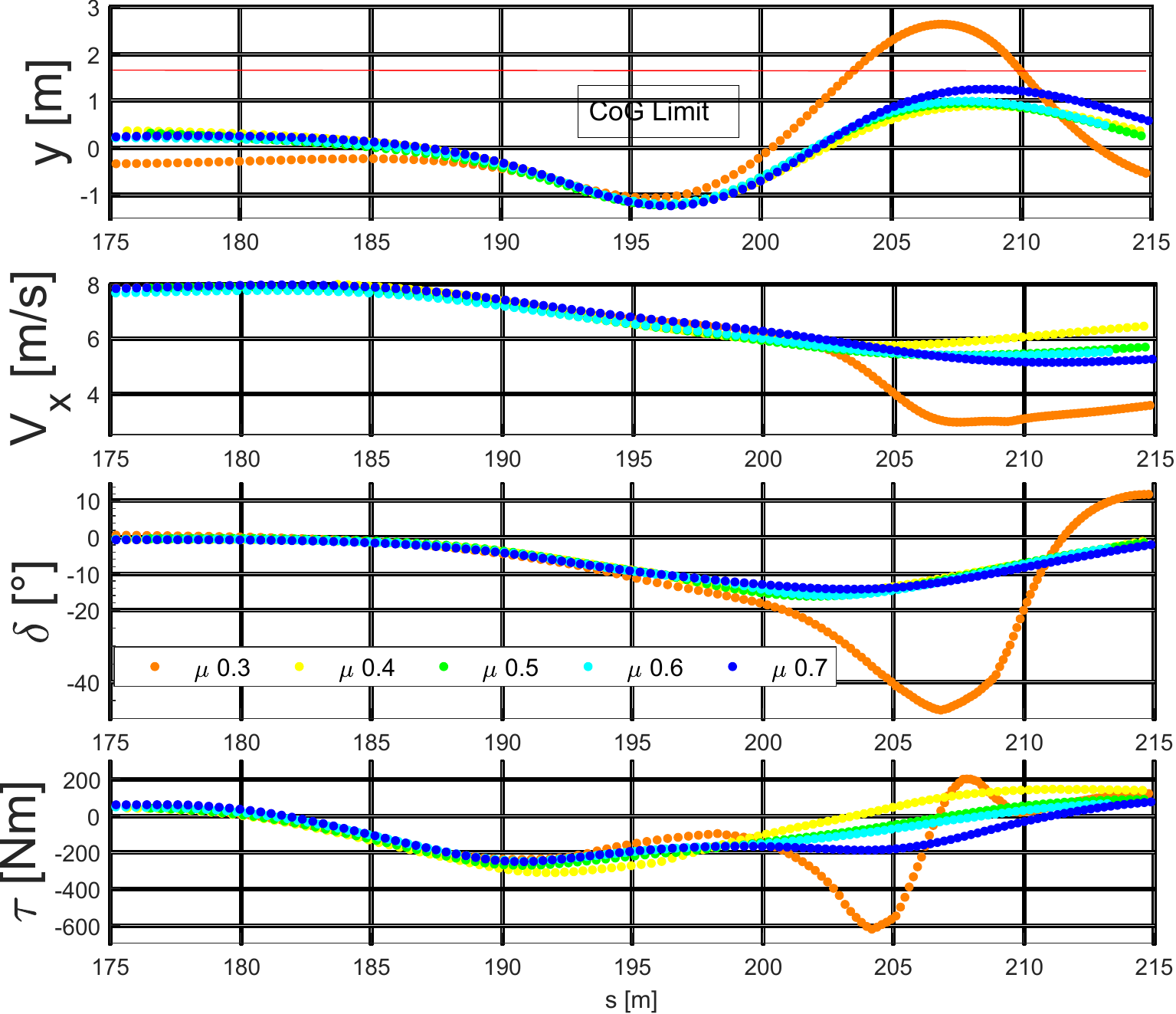}} 
	\caption[Simulation using different adherence coefficients]{Simulation using different adherence coefficients (reference speed equal to $8 m/s$)}
	\label{fig:res_10t}
\end{figure}

Results are shown in fig.\ref{fig:res_10t}, where time evolution of lateral displacement, longitudinal speed, steering angle and applied torque. In detail, fig.\ref{fig:res_10} represents the first turn. Simulation at $\mu=0.2$ (snow) presents a loose of vehicle's stability as well shown by lateral displacement though opposite steering command applied, while heavy rain condition ($\mu=0.3$) cause a violation of road borders constraints. Sharp turn in fig.\ref{fig:res_11} confirmed the bad behavior shown in turn 1 empathizing constraints violation in heavy rain scenario as well shown by lateral displacement and an uncomfortable variation of applied torque.
The decisional algorithm behave well in the range $\mu\in[0.4,0.7]$ showing a good rejection to parameter estimation errors.


\subsubsection{Stationary vehicle}
\begin{figure}[htp]\centering
	\includegraphics[width=0.95\columnwidth]{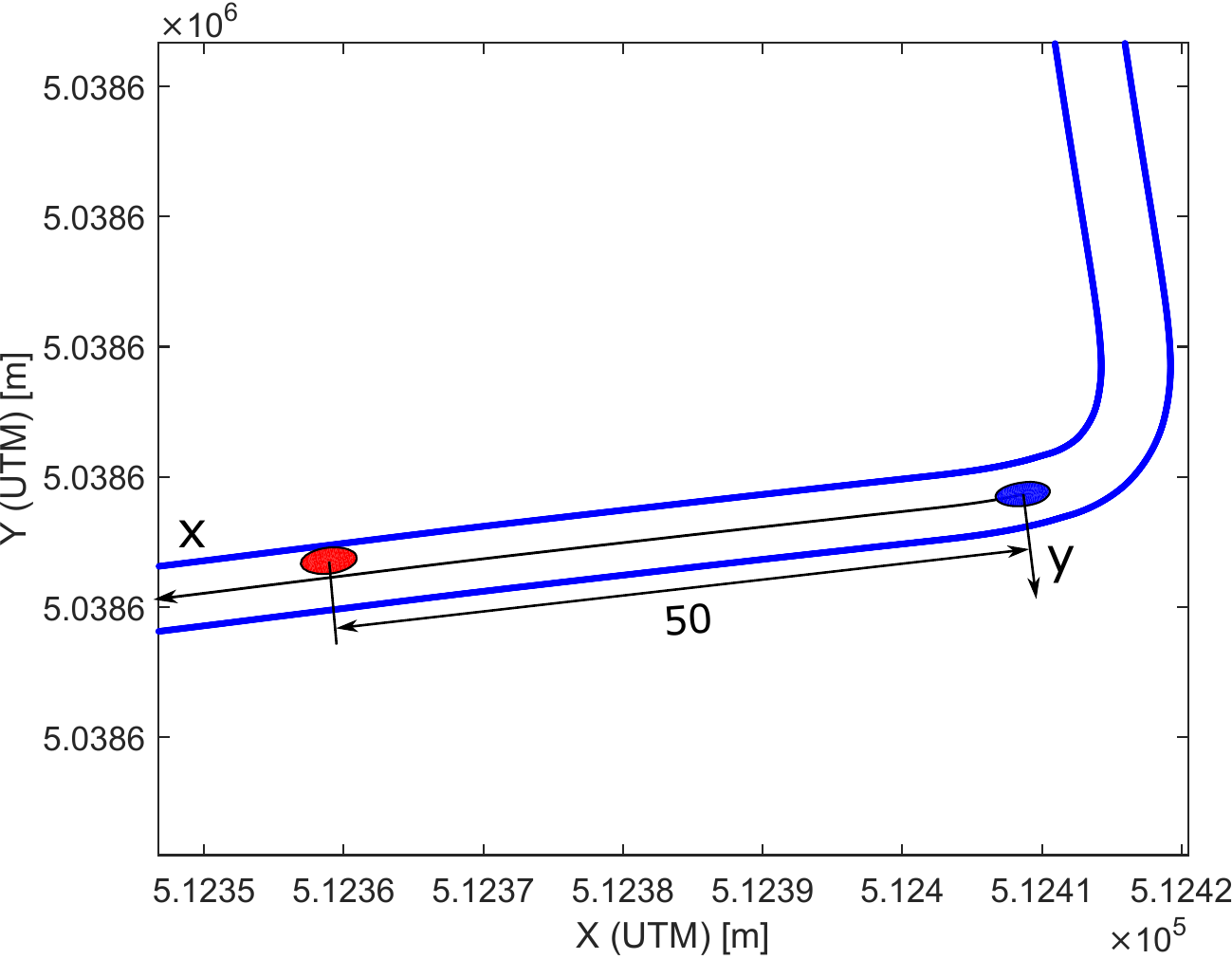}
	\caption[Stationary vehicle scenario]{Stationary vehicle scenario: red ellipse is a parked vehicle on the right side of the road, while blue ellipse is controlled vehicle at the beginning of the simulation}
	\label{fig:res_5}
\end{figure}

The simulations proposed in this section involve a dangerous urban situation where a vehicle is parked on the right side of the road (i.e. picking up a pedestrian or waiting for a parking spot). In fig. \ref{fig:res_5} the scenario considered is shown. 
Obstacle vehicle is simplified by an ellipse whose minor and major semi-axis are $1m$ and $2m$ respectively and it's lateral position is $-1.3m$ respect to the centerline of the road.
The scenario presented is investigated by considering four increasing reference speeds. In detail $V_{x\_ref}$ is set to  $[4,8,12,18]$  $m/s$.

\begin{figure}
	\centering
	\subfloat[][Fixed obstacle\label{fig:res_12a}]
	{\includegraphics[width=0.95\columnwidth]{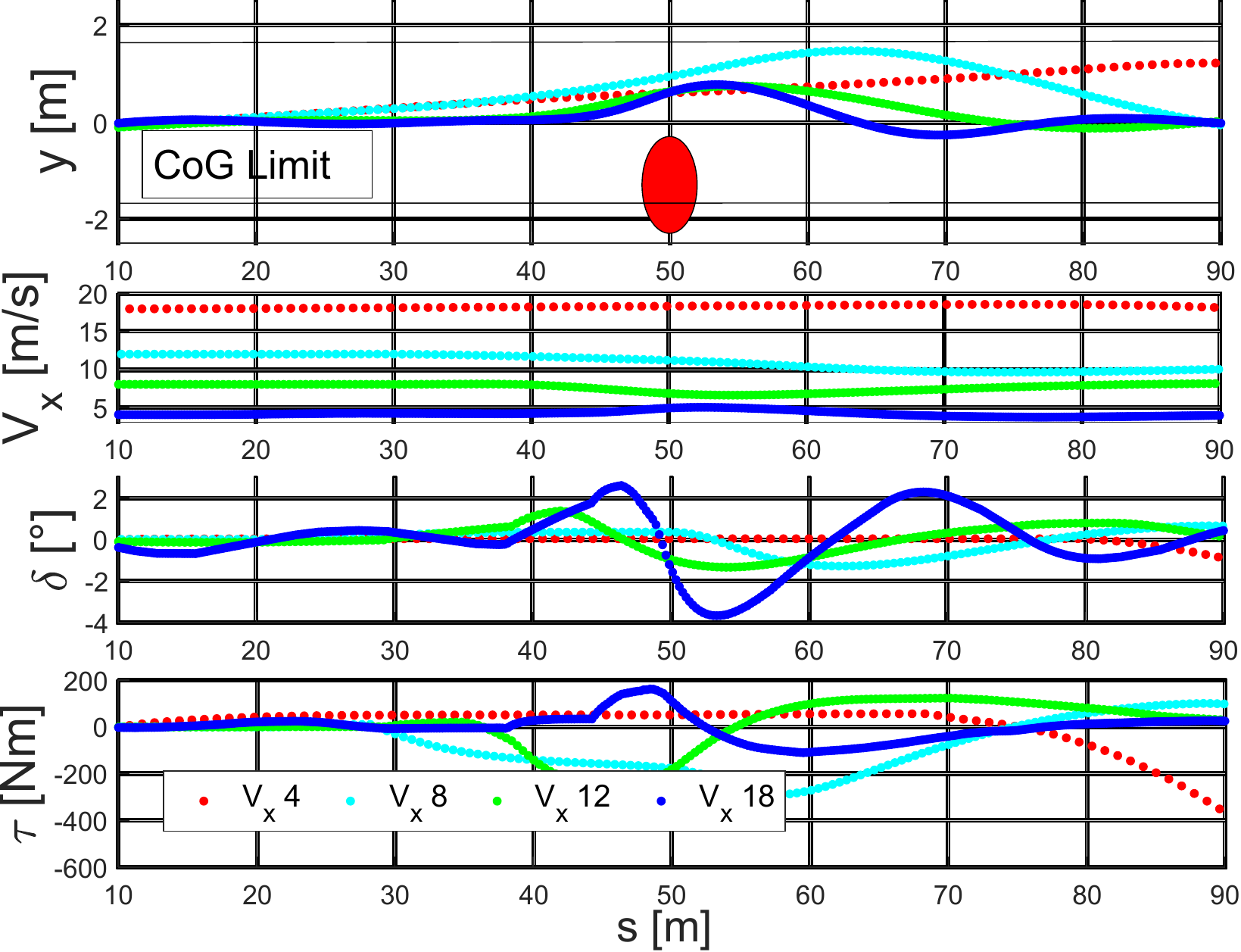}} \\
	\subfloat[][Fixed obstacle + turn 1\label{fig:res_12b}]
	{\includegraphics[width=0.95\columnwidth]{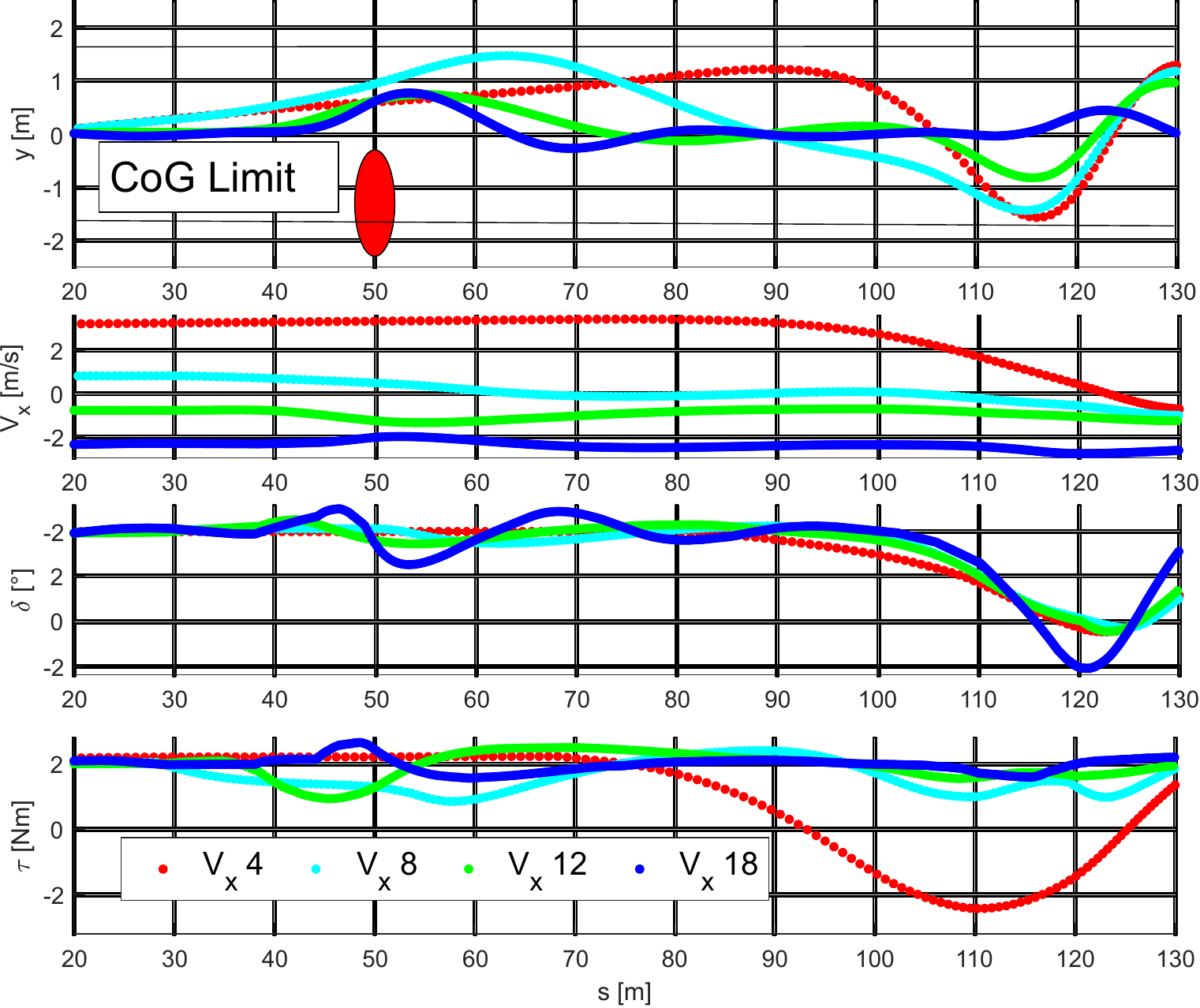}} 
	\caption[Fixed obstacle avoidance + turn 1 at different speed]{Fixed obstacle avoidance + turn 1 at different speed: a comparison between simulation results obtained by different reference velocities of the vehicle approaching a static obstacle and a sharp turn}
	\label{fig:res_12}
\end{figure}

Fig.\ref{fig:res_12} presents an extensive comparison between simulation results. In particular lateral displacement $y$ satisfy the limits in all cases. In detail, as noticed also in section \ref{sec:res_complete}, longer spatial predictive horizon and $V_y,\dot{\delta},\dot{\tau}$ cost terms generate a smoother evasive maneuver for high speed simulations as shown in fig.\ref{fig:res_12a}. Fig.\ref{fig:res_12b} also considers the subsequent right turn. Even if all simulation succeed, high speed simulations ($12,18$) require a sudden speed reduction approaching the corner as expected.

\subsubsection{Multiple moving obstacles}
\begin{figure}[htp]\centering
	\includegraphics[width=0.95\columnwidth]{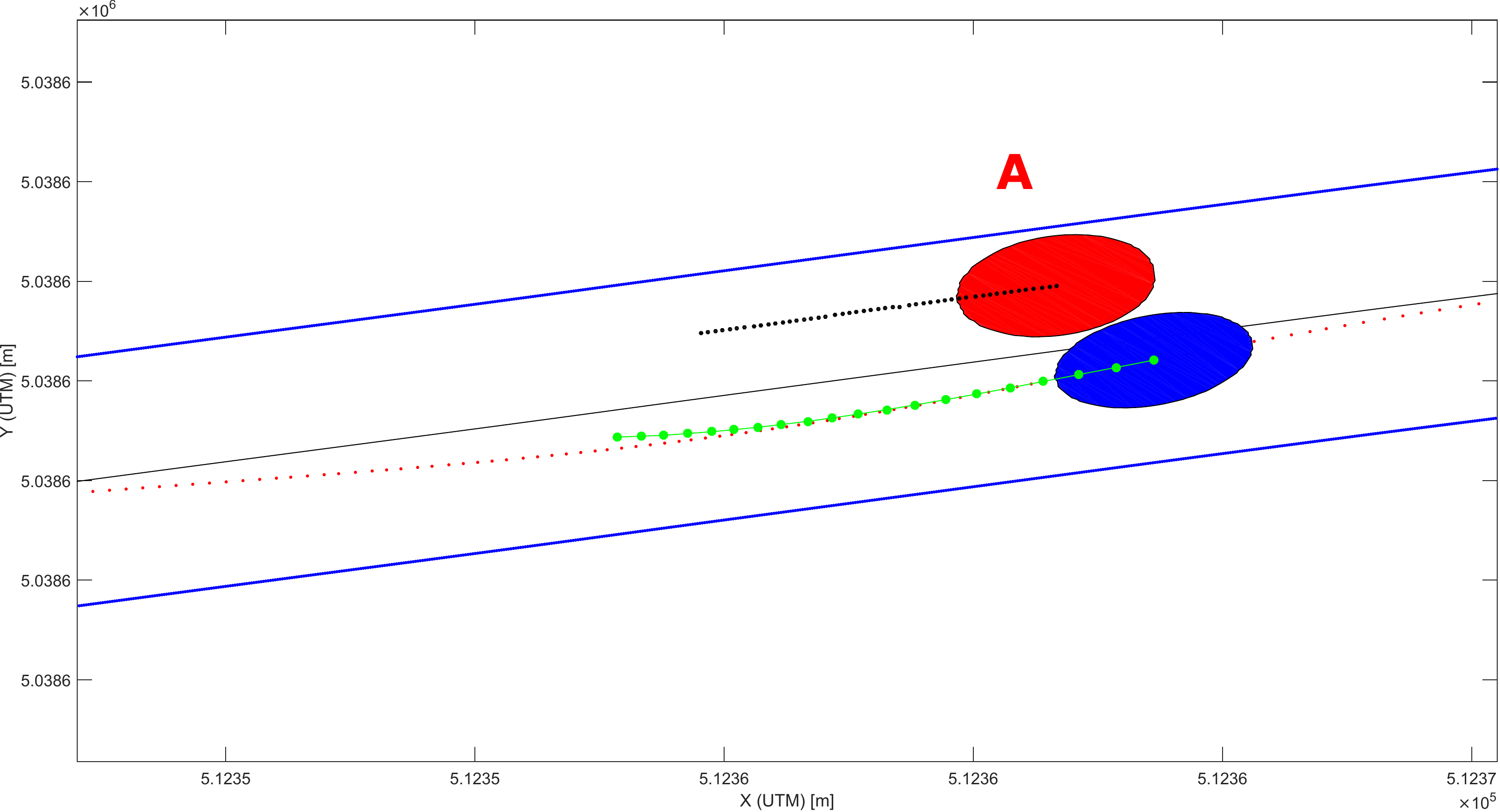}
	\caption[Moving obstacles's scenario - frame 1]{Moving obstacles's scenario - frame 1: overtaking maneuver in progress. Green line represents planned trajectory over the predictive horizon, red line is the closed-loop trajectory performed and the black lines are estimated obstacles' motion (reference speed equal to $8 m/s$)}
	\label{fig:res_13_a}
\end{figure}

\begin{figure}[htp]\centering
	\includegraphics[width=0.95\columnwidth]{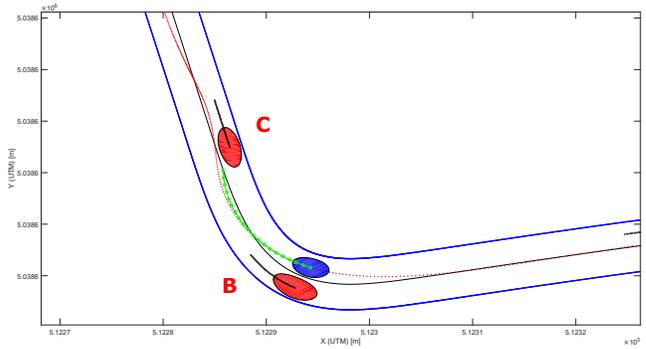}
	\caption[Moving obstacles's scenario - frame 2]{Moving obstacles's scenario - frame 2:Overtaking and turning}
	\label{fig:res_13_b}
\end{figure}

\begin{figure}[htp]\centering
	\includegraphics[width=0.95\columnwidth]{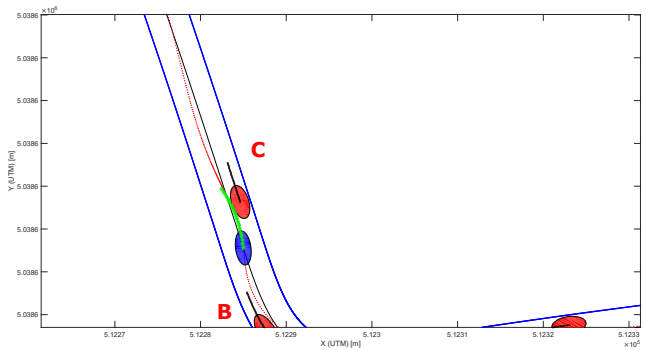}
	\caption[Moving obstacles's scenario - frame 3]{Moving obstacles's scenario - frame 3: slow down and additional evasive maneuver planning.}
	\label{fig:res_13_d}
\end{figure}

\begin{figure}[htp]\centering
	\includegraphics[width=0.95\columnwidth]{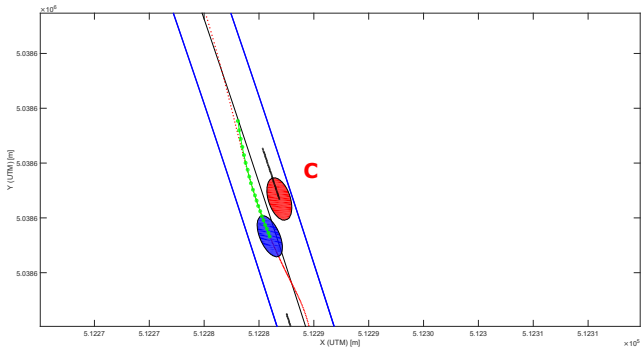}
	\caption[Moving obstacles's scenario - frame 4]{Moving obstacles's scenario - frame 4: overtaking and speed increase}
	\label{fig:res_13_e}
\end{figure}

Last simulation proposed involves a complex urban traffic scenario with multiple obstacles moving on the road. 
\begin{figure}[htp]\centering
	\includegraphics[width=0.95\columnwidth]{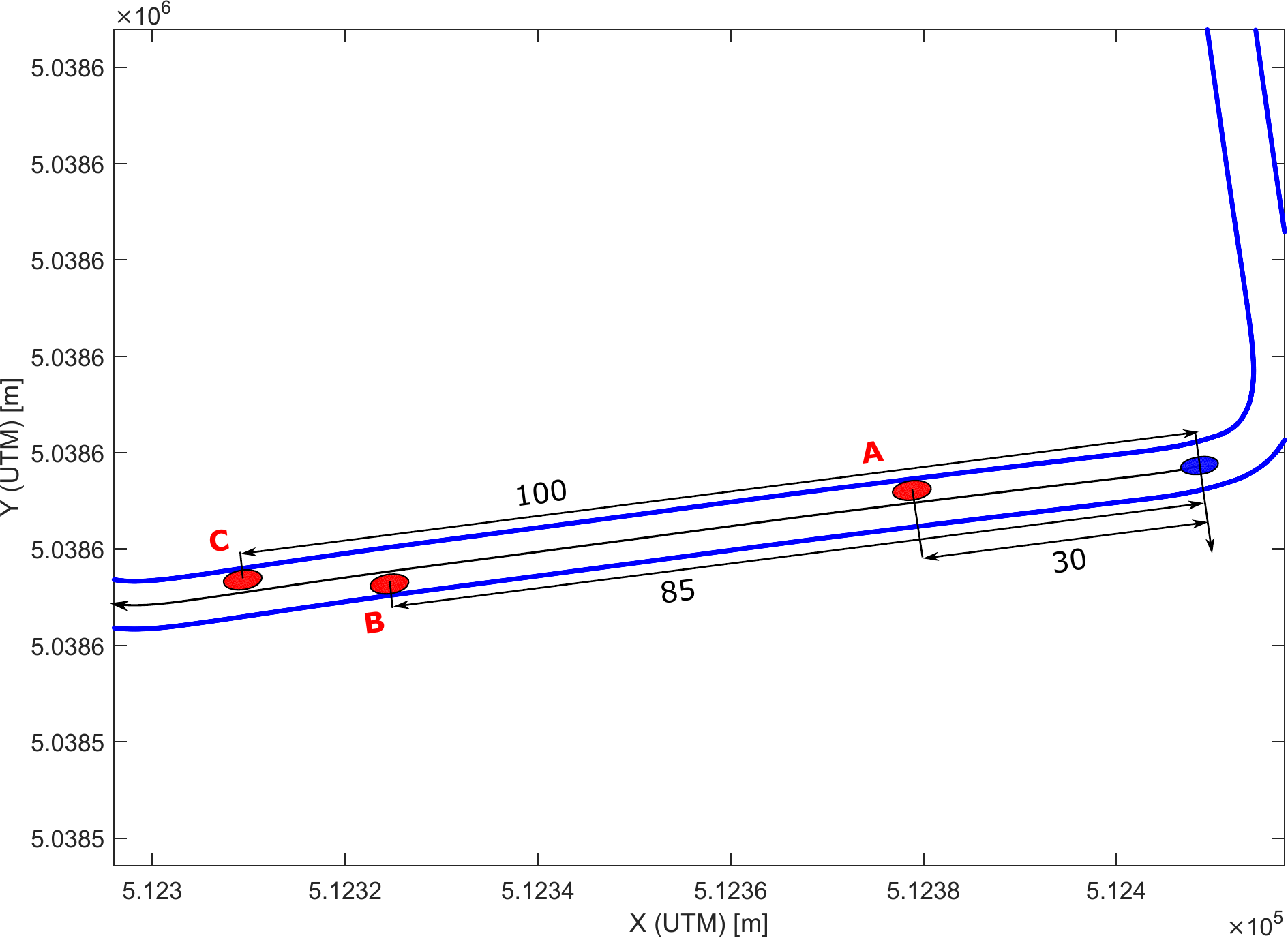}
	\caption[Multiple moving obstacles' scenario]{Multiple moving obstacles' scenario: where the obstacles considered are vehicle \emph{A} (driving on right side), vehicle \emph{B} (driving on left side) and vehicle \emph{C} (driving on right side)}
	\label{fig:res_6}
\end{figure}
As shown in fig. \ref{fig:res_6}, vehicle \emph{A} is driving on right side ($-1.3m$) at $3m/s$, vehicle \emph{B} is driving on left side ($1.3m$) at $2m/s$ and finally vehicle \emph{C} is driving on right side ($-1.3m$) at $3m/s$.
The simulation is performed imposing a reference speed equal to $8 m/s$.
The simulation is presented by means of a sequence of figures representing different time frames (shown in figs. \ref{fig:res_13_a,fig:res_13_b,fig:res_13_d, fig:res_13_e}) in order to clarify controller behavior. In all figures blue ellipse represents controlled vehicle, green line is the optimal trajectory planned by the controller over the predictive horizon considered and red line is the closed-loop trajectory followed by the vehicle during the simulation. Red ellipses represent obstacles and the black lines connected to them are expected obstacle motion of obstacles estimated by means of linear extrapolation as explained in section \ref{sec:ref_numsim}. 
it's assumed that controlled vehicle is allowed to overtake regardless its position on the road (even during a turn) or according to road rules (overtaking on left/right). This is done as a stress test of the algorithm: in a rel driving situation higher driving tasks in the hierarchical scheme in fig.\ref{fig:decisional_scheme}, would prevent these situations. Moreover obstacle's actual state are considered always known (regardless if they are visible or not by on-board sensors).
In fig. \ref{fig:res_13_a}, controlled vehicle is overtaking vehicle \emph{A}. 
When approaching the first corner, as shown in fig.\ref{fig:res_13_b}, a cornering maneuver is defined considering  (and correctly avoiding) vehicle \emph{B} that is cornering too on left side of the road. 
Concluded that maneuver, a slow vehicle (\emph{C}) suddenly appears in the predictive horizon of the replanner forcing it to slow down and plan another overtaking maneuver as visible in fig.\ref{fig:res_13_d}. 
Finally the resulting overtaking maneuver is shown in fig.\ref{fig:res_13_e} where predictive horizon spatial length implies that vehicle is expected to speed up.

\section{Conclusion}\label{sec:CONC}
In this paper a NMPC trajectory planner for autonomous vehicles based on a direct approach has been presented. Thanks to the proposed modified slip calculation, the dynamic single-track vehicle model can be employed at both high and low velocities. This allows to have one single model that presents a kinematic-like behaviour at low velocities while keeping the standard dynamic single-track vehicle model behaviour at higher speeds. Obstacles' uncertainties have been taken into account through the implementation of coupled soft and hard constraints. The correlation between the simulated and perceived spatial horizon lengths has been enforced with a spatial dependent velocity profile that has been imposed as an extra inequality constraint. The numerical solution is carried out using ACADO toolkit, coupled with the QP solver qpOASES. The trajectory planner performances have been checked in simulation, applying the controller to a realistic nonlinear multibody model in CarMaker environment. Two significant driving scenarios have been reported assessing the trajectory planner performances in terms of feasibility of the generated trajectories and passengers' comfort. The analysis of the computational cost has confirmed that the proposed trajectory planner can be implemented in real-time.

In this paper a novel decisional algorithm for trajectory planning \& tracking is proposed.
In detail the algorithm is expected to ensure: a real-time feasible implementation, a sufficient robustness of the solution provided respect to modeling errors and being able to deal with multiple and moving obstacles in urban-like scenarios.

The algorithm consists mainly in a optimal constrained problem formulated as a MPC problem and numerically solved by means of a novel genetic algorithm approach. 
In particular, thanks to GA implementation for numerical optimization compared to other approaches in literature, the proposed algorithm prevents the solution to get stuck in local minima and the specific mathematical formulation proposed bound computational cost and generates continuous control commands at the same time. Especially, the specific vehicle model adopted and the choice of optimization variables force the evasive maneuver calculated over the predictive horizon of MPC to present continuous acceleration and jerk, while trajectory computed by the closed loop presents continuous acceleration in time. on the other hand, this approach limits the search of a solution to a suboptimal subset (due to hypothesis on control variables as linear equation). This assumption is considered reasonable thanks to the short predictive horizon and the short sampling time of the controller.

Numerical simulations of a typical urban area (made of sharp curves and narrow lanes) are shown to test the algorithm
Computational time measured during the simulations confirms the feasibility of a real-time implementation of the controller for working frequency up to $20Hz$, as well as sufficient robustness respect to modeling errors and its capability to deal with multiple and moving obstacles in urban-like scenarios is shown.



\section*{Acknowledgment}
The authors would like to thank Sineco s.p.a for the possibility to use an instrumented vehicle for high accuracy measuments. The Italian Ministry of Education, University and Research is acknowledged for the support provided through the Project "Department of Excellence LIS4.0 - Lightweight and Smart Structures for Industry 4.0”.
%
%
%



%
%

\bibliographystyle{IEEEtran}
\bibliography{Arrigoni_bib_AV} 
%
\end{document}